\documentclass[lettersize,journal]{IEEEtran}
\usepackage{amsmath,amsfonts}
\usepackage{algorithmic}
\usepackage{algorithm}
\usepackage{array}
\usepackage[caption=false,font=normalsize,labelfont=sf,textfont=sf]{subfig}
\usepackage{textcomp}
\usepackage{stfloats}
\usepackage{url}
\usepackage{verbatim}
\usepackage{graphicx}
\usepackage{cite}
\usepackage{hyperref}
\usepackage{multirow}
\usepackage{microtype}
\usepackage{graphicx}
\usepackage{booktabs}
\usepackage{bm}
\usepackage{tabularx}
\usepackage{amsmath}
\usepackage{amssymb}
\usepackage{mathtools}
\usepackage{amsthm}
\usepackage{gensymb}

\usepackage{hyperref}
\usepackage{balance}
\usepackage{diagbox}
\usepackage{color}
\usepackage[capitalize,noabbrev]{cleveref}
\usepackage[textsize=tiny]{todonotes}

\begin{document}

\title{Toward Robust and Accurate 
 Adversarial Camouflage Generation against Vehicle Detectors}

\author{Jiawei Zhou, Linye Lyu, Daojing He, and Yu Li
\thanks{Manuscript received October 18, 2024; revised at May 8, 2025; accepted July 20, 2025. \textit{(Corresponding author: Yu Li)}}
\thanks{Jiawei Zhou, Linye Lyu, Daojing He are with School of Computer Science and Technology, Harbin Institute of Technology, Shenzhen 518055, Guangdong, P.R. China (email: 2798589537@qq.com; lyulinye@gmail.com; hedaojinghit@163.com). }
\thanks{Yu Li is with College of Integrated Circuits, Zhejiang University, Hangzhou 310000, Zhejiang, P.R. China (e-mail: yu.li.sallylee@gmail.com).}
}

\markboth{Journal of \LaTeX\ Class Files,~Vol.~14, No.~8, August~2021}%
{Shell \MakeLowercase{\textit{et al.}}: A Sample Article Using IEEEtran.cls for IEEE Journals}

\IEEEpubid{0000--0000/00\$00.00~\copyright~2021 IEEE}

\maketitle

\begin{abstract}
Adversarial camouflage is a widely used physical attack against vehicle detectors for its superiority in multi-view attack performance. One promising approach involves using differentiable neural renderers to facilitate adversarial camouflage optimization through gradient back-propagation. However, existing methods often struggle to capture environmental characteristics during the rendering process or produce adversarial textures that can precisely map to the target vehicle. 
Moreover, these approaches neglect diverse weather conditions, reducing the efficacy of generated camouflage across varying weather scenarios. To tackle these challenges, we propose a robust and accurate camouflage generation method, namely RAUCA. The core of RAUCA is a novel neural rendering component, End-to-End Neural Renderer Plus (E2E-NRP), which can accurately optimize and project vehicle textures and render images with environmental characteristics such as lighting and weather.
In addition, we integrate a multi-weather dataset for camouflage generation, leveraging the E2E-NRP to enhance the attack robustness. Experimental results on six popular object detectors show that RAUCA-final outperforms existing methods in both simulation and real-world settings. 
\end{abstract}


\section{Introduction}

\IEEEPARstart{D}{eep} Neural Networks (DNNs) have achieved remarkable performance in many real-world applications such as face recognition and autonomous vehicles \cite{krizhevsky2012imagenet,wang2023yolov7,chai2021deep}. However, DNNs suffer from adversarial examples \cite{szegedy2013intriguing}. For instance, in the task of vehicle detection,  adversarial inputs can deceive detection models, leading to incorrect detection of the surrounding vehicles, which poses a severe threat to the safety of autonomous vehicles.

\begin{figure}[ht]
\vfill
\begin{center}
\centerline{\includegraphics[width=\columnwidth]{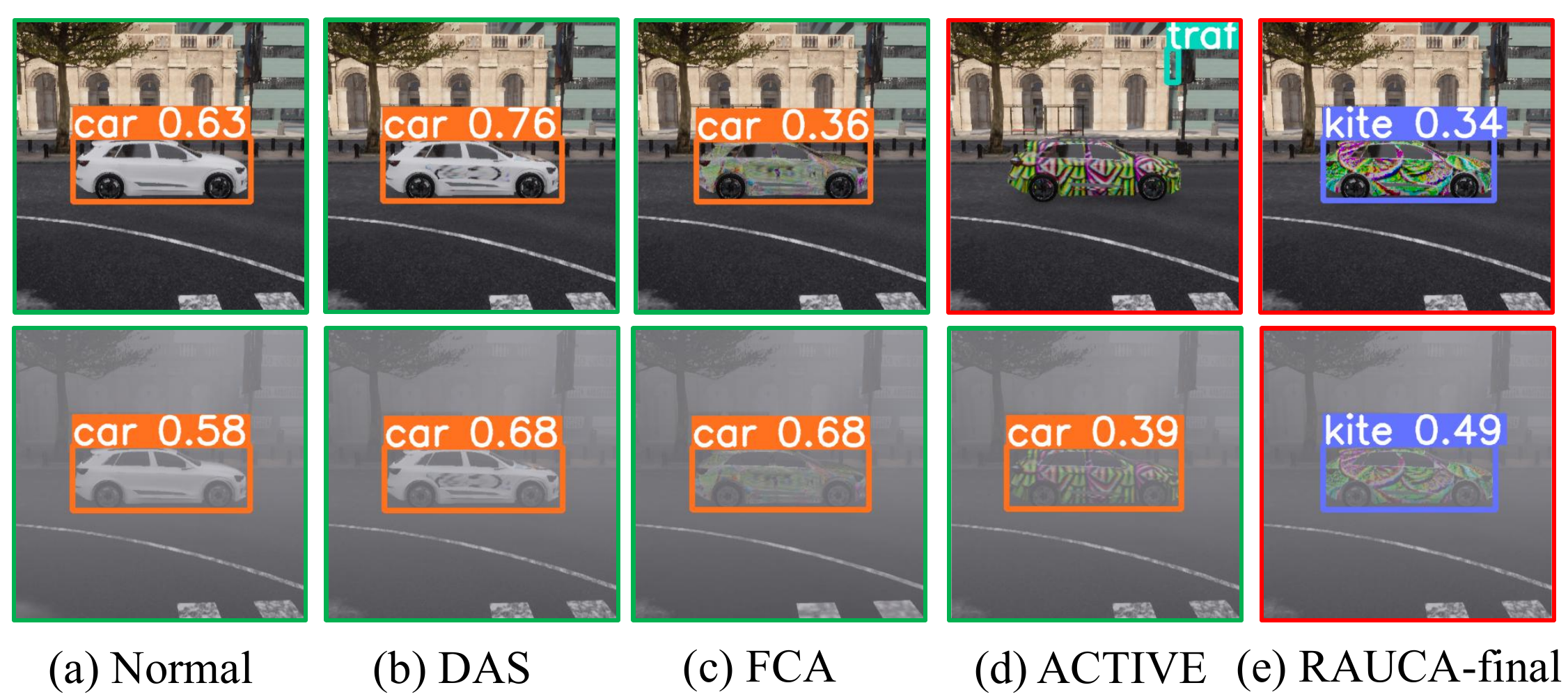}}
\caption{ Comparison of different adversarial camouflage under sunny (first row) and foggy (second row) environments, where only our method succeeds in both cases. (a) A car with normal texture. (b) DAS \cite{wang2021dual}. (c) and (d) are top-performed methods FCA \cite{wang2022fca} and ACTIVE \cite{Suryanto_2023_ICCV}, respectively. (e) Our method RAUCA-final. 
}
\label{introduction}
\end{center}
\vfill
\end{figure}
Adversarial examples can be categorized into digital and physical adversarial types. Digital adversarial examples involve introducing pixel-level perturbations to the model's digital input, while physical ones manipulate real-world objects or surroundings, indirectly influencing the model inputs \cite{shapeshifter,Crime,Understanding}. 
Physical adversarial examples are generally deemed more practical, as gaining direct access to the model's input often necessitates system authentication. 
However, they are inherently more challenging as they must prove effective in complex physical environments, including various viewing perspectives, spatial distances, and lighting/weather conditions.

This paper focuses on physical adversarial examples against vehicle (e.g., car) detection models, given their wide adoption in autonomous driving scenarios where safety is of great importance. To ensure the attack effectiveness across various viewing angles, current methods prefer generating adversarial camouflage capable of covering the entire surface of the vehicle \cite{wang2021dual,wang2022fca,Suryanto_2022_CVPR,Suryanto_2023_ICCV}. 
Top-performing methods achieve this by leveraging a differential neural renderer. This renderer maps the 3D vehicle and its texture to 2D images, establishing a differentiable path between the 3D vehicle and the vehicle detection models. Thus, the texture can be optimized through gradient back-propagation for effective camouflage generation.

\IEEEpubidadjcol


\begin{table}[t]
\caption{Comparison of proposed and existing physical camouflage attack methods.}
\label{method compare}
\begin{center}
\begin{small}
\begin{sc}
\resizebox{1\columnwidth}{!}{
\begin{tabular}{llcccc}
\toprule
Method Type & Method & Acc. Render & Real. Render& Multi-weather & E2E
     \\
\midrule
\multirow{2}{*}{World-aligned} 
 & DTA          & $\times$& $\checkmark$ & $\times$ & $\checkmark$ \\
& ACTIVE       & $\times$ &  $\checkmark$ & $\times$ & $\checkmark$ \\
\midrule
\multirow{4}{*}{UV-map} 
 & DAS          & $\checkmark$ & $\times$ & $\times$ & $\times$ \\
 & FCA          & $\checkmark$ & $\times$ & $\times$ & $\times$ \\
 & RAUCA-Base        & $\checkmark$ & $\checkmark$ & $\checkmark$ & $\times$ \\
 & RAUCA-Final  & $\checkmark$ & $\checkmark$ & $\checkmark$ & $\checkmark$ \\
\bottomrule
\end{tabular}
}
\end{sc}
\end{small}
\end{center}
\vskip -0.1in
\quad \quad \quad\begin{scriptsize}
\begin{flushleft}
\textbf{Notes:}\\
(1) Acc. Render: whether the renderer accurately renders camouflage patterns to ensure consistency in the projected texture on the vehicle during generation and deployment.  \\
(2) Real. Render: whether the renderer realistically renders environments for consistent foreground-background rendering \\
(3) Multi-weather:  whether multi-weather robustness is considered during optimization  \\
(4) E2E:  whether camouflage textures are optimized end-to-end without requiring post-processing for deployment.
\end{flushleft}
\end{scriptsize}
\end{table}

 In the literature, there are two ways to generate camouflage with a neural renderer: one way is to optimize a 2D square texture pattern and project it onto the vehicle repeatedly, referred to as world-align-based methods; the other is to optimize the 3D texture of the vehicle in the form of UV-maps, referred to as UV-map-based methods. However, both approaches currently suffer from certain issues. The world-align-based methods \cite{Suryanto_2022_CVPR,Suryanto_2023_ICCV} cannot guarantee the texture pattern projected onto the car in the same manner in texture generation and deployment, leading to differences in adversarial camouflage between generation and deployment. In contrast, the UV-map-based methods \cite{wang2021dual,wang2022fca} can address this inaccuracy, as the optimized UV map can be accurately deployed on the vehicle. While promising, however, they cannot optimize the UV map in an end-to-end manner. Instead, they optimize a facet-based texture tensor, which requires a non-differentiable post-processing step to convert it into a deployable UV map. Moreover, the renderer in these UV-map-based methods cannot render sophisticated environment characteristics such as light and weather, leading to limited performance. Please note that all the above methods fail to consider the effectiveness of the camouflage under various weather conditions. As shown in Figure \ref{introduction}, the camouflage generated by the top-performed method ACTIVE \cite{Suryanto_2023_ICCV} fails to attack the detector in a foggy scene.

 To address the above issues, we develop a novel adversarial camouflage generation framework against vehicle detectors. Our key insight is that, for successful physical attacks, the generated camouflage must accurately map onto the vehicle. Additionally, the camouflage needs to be robust under different environmental conditions. Achieving the first goal necessitates a genuine end-to-end optimization of the UV map. The second goal requires a dataset encompassing ample environmental effects and also requires that these effects can be successfully utilized for camouflage optimization. To achieve the first goal, we choose a UV-map-based neural renderer \cite{kato2018neural} as our base renderer and enhance the renderer's capability to backpropagate gradients to the UV map. However, the current sampling method from the UV map to the facet-based texture tensor causes numerous points on the UV map unoptimized. To address this issue, we then refine the sampling method, allowing all of the points on the UV map to be optimized, thus achieving genuine end-to-end optimization.
To achieve the second goal, we then integrate the renderer with an Environment Feature Extractor (EFE) network to render environmental information, allowing environmental effects to be effectively utilized in camouflage optimization. Furthermore, we augment the dataset used for camouflage generation to include diverse weather conditions. This enhanced dataset, combined with the refined renderer, facilitates the generation of robust camouflage.


Our contributions can be summarized as follows:
 \begin{itemize}
\item We present the Robust and Accurate UV-map-based Camouflage Attack (RAUCA), a framework for generating physical adversarial camouflage against vehicle detectors. It enhances the effectiveness and robustness of the adversarial camouflage through a novel rendering component and a multi-weather dataset.
\item  We propose End-to-End Neural Renderer Plus (E2E-NRP), a neural renderer capable of accurate and realistic rendering for end-to-end camouflage optimization. To facilitate accurate camouflage generation, we leverage the UV-map-based neural renderer\cite{kato2018neural} as the base renderer. We then adapt it to achieve realistic environment information rendering and refer to this enhanced model as Neural Renderer Plus (NRP). Furthermore, we enable NRP to be capable of genuine end-to-end optimization and call this advanced renderer End-to-End Neural Renderer Plus (E2E-NRP).

\item We construct a large-scale multi-weather dataset, containing a wide range of environmental variations (e.g., lighting, rain, fog) and diverse camera viewpoints. Our experiments show that using this dataset substantially enhances the attack robustness.

\item We accelerate the training methodology of the environment feature extractor (EFE) within the E2E-NRP framework. We leverage the construction of an offline-rendered vehicle image set and shared EFE inference strategy to significantly reduce the computations of rendering process and EFE inference, thereby enhancing the overall training efficiency of EFE.


\item We introduce a pretrained EFE to accelerate adaptation. To generalize the RAUCA framework to previously unseen vehicle models, we introduce a pretrained Environment-Facing Encoder (EFE) that has been trained across multiple vehicle types. This design enables the system to rapidly adapt to new vehicle appearances with minimal fine-tuning, significantly reducing the training overhead required for deployment.

\end{itemize}

Our extensive studies demonstrate that our method outperforms the current state-of-the-art method by around 14\% in car detection performance (e.g., AP@0.5). Our attack achieves strong robustness and effectiveness, excelling in multi-weather, multi-view, and multi-distance conditions. We also demonstrate that our method is effective in the physical world. Our code is available at: \href{https://github.com/SeRAlab/RAUCA-E2E}{https://github.com/SeRAlab/RAUCA-E2E}.

 This work extends our ICML paper \cite{RAUCA} in several aspects. 1) We refine the neural renderer to facilitate genuine end-to-end optimization. It directly optimizes the UV map, thereby avoiding attack performance decline associated with the post-processing from the final optimized facet-based texture tensor to the UV map. Moreover, we propose a novel sampling method from the UV map to the facet-based texture tensor to allow all points on the UV map to be optimized, thus generating camouflage with better attack effects. 2) We enhance the multi-weather dataset by incorporating more weather combinations, utilizing a broader range of camera perspectives in contrast to the previously fixed 128 viewpoints, and modifying the vehicle paint material to align with our physical camouflage implementation material. 3) We enhance the training methodology of the EFE network to mitigate the computations of rendering process and EFE inference, significantly reducing overall training time. 4) We train the EFE component with multiple objects to obtain pre-trained EFE. It decreases EFE training time for unseen vehicles, thus enhancing the adaptability of our attack framework. 5) We follow \cite{Suryanto_2023_ICCV} to add the Random Output Augmentation component into the attack framework to boost the robustness of our generated camouflage.

\section{Related work}
\textbf{Physical Adversarial Attack}: Physical adversarial attacks need to consider the robustness of the attacks because the objects and the environment are not constant. The Expectation over Transformation (EoT) \cite{athalye2018synthesizing} is a prime method of generating robust adversarial examples under various transformations, such as lighting conditions, viewing distances, angles, and background scenes. As a result, many adversarial camouflage methods \cite{zhang2019camou, wu2020physical,wang2021dual,wang2022fca, Suryanto_2022_CVPR, Suryanto_2023_ICCV} employ EoT-based algorithms to enhance their attack robustness in the real world scenarios.


\textbf{Adversarial Camouflage}:
For self-driving cars, precise detection of surrounding vehicles is a critical safety requirement. Consequently, there has been a growing interest in developing adversarial vehicle camouflage to evade vehicle detection systems. Current research works mostly leverage a 3D simulation environment to obtain 2D rendered vehicle images with various transformations to obtain robust adversarial camouflage. Early research on adversarial camouflage for vehicles was mostly done using black-box methods because the rendering process in the simulation environment was non-differentiable. \cite{zhang2019camou} first conducts experiments in the 3D space to generate adversarial camouflage for cars. They propose CAMOU, a method to train an approximate gradient network to mimic the behavior of both rendering and detection of the camouflage vehicles. Then, they can optimize the adversarial texture using this network. Meanwhile, \cite{wu2020physical} proposes an adversarial camouflage generation framework based on a genetic algorithm to search an adversarial texture pattern. Then, they repeat the optimized texture pattern to build the 3D texture that covers the full-body vehicle. 

Recent methods introduce neural renderers, which enable differentiable rendering. With this technique, the adversarial texture can be optimized via gradient back-propagation. Two primary methods are employed in adversarial camouflage generation with neural renderer. One is to optimize a 3D model texture, which we refer to as UV-map-based camouflage methods; for example, \cite{wang2021dual} propose a Dual Attention Suppression (DAS) attack, which minimizes the model attention and human attention on the camouflaged vehicle. Besides, \cite{wang2022fca} propose the Full-coverage Camouflage Attack (FCA), which optimizes full-body surfaces of the vehicle in multi-view scenarios. The other is to optimize a 2D square texture pattern and then project it repeatedly to the target vehicle surface, which we refer to as world-align-based camouflage methods. \cite{Suryanto_2022_CVPR} present the Differentiable Transformer Attack (DTA), which proposes a differentiable renderer that can express physical and realistic characteristics (e.g., shadow). ACTIVE, proposed by \cite{Suryanto_2023_ICCV}, introduces a texture mapping technique that utilizes depth images and improves the naturalness of the camouflage by using larger texture resolution and background colors. 

While prior neural-renderer-based methods have achieved impressive attack success rates, they typically struggle to capture the environmental characteristics such as shadow or produce adversarial camouflage that can precisely map to the target vehicle, resulting in sub-optimal camouflage. Moreover, these methods frequently overlook diverse weather conditions during the generation of adversarial camouflage, hindering the effectiveness of the camouflage in varying weather environments. Recently, in a different task—attacks on monocular depth estimation—recent work \cite{zheng2024physical} has incorporated weather factors by adding noise and perturbations to the rendered vehicle foreground during the camouflage generation process. Nevertheless, this approach has a key limitation: the noise is applied solely to the foreground, resulting in weather inconsistencies between the foreground and background, which in turn reduces the realism of the synthesized images.

Table \ref{method compare} compares existing approaches with the proposed method across various criteria. As shown, our method achieves superior performance in all evaluation dimensions compared to previous approaches.

\section{Method}
\begin{figure*}[t]
\vfill
\begin{center}
\centerline{\includegraphics[width=\textwidth]{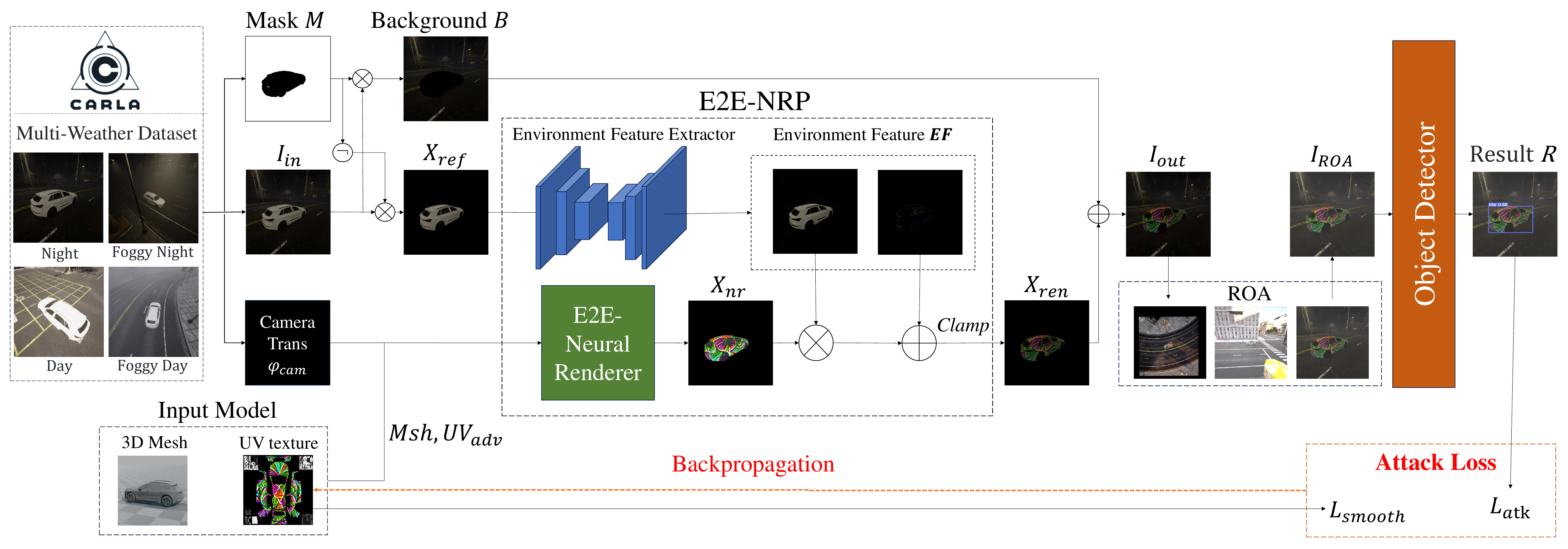}}
\caption{The overview of RAUCA. First, we create a multi-weather dataset using CARLA, which includes car images, corresponding mask images, and camera transformation sets. Then, the car images are segmented using the mask images to obtain the foreground car and background images. The foreground car image, the 3D model, and the camera transformation are passed through the E2E-NRP rendering component for rendering. The rendered image is then seamlessly integrated with the background. After a series of random output augmentation, the image is fed into the object detector. Finally, we optimize the adversarial camouflage through back-propagation with our devised loss function computed from the output of the object detector.}
\label{pipeline}
\end{center}
\vfill
\end{figure*}

In this section, we present the overview of our framework for generating adversarial camouflage. Subsequently, we provide an in-depth explanation of the essential components comprising our attack framework, RAUCA.

\subsection{Overview}

Figure \ref{pipeline} shows our entire framework for adversarial camouflage generation. Firstly, we modify the weather parameters in the simulation environment to
obtain a multi-weather vehicle dataset (\(I_{in}\), \(Y\), \(\Phi_{cam}\), \(\Phi_{M}\), 
\(M\)), where \(I_{in}\) is the original  input images, \(Y\) is the
ground truth labels, \(\Phi_{cam}\) is the camera transformation parameters
 (position and angle) for viewing the car, \(\Phi_{M}\) is the car locations within the simulation environment and \(M\) is the binary masks of \(I_{in}\), where the target vehicle areas are set to 0. With \(I_{in}\) and
\(M\), we can use:
\begin{gather}
X_{r e f}=I_{i n} \cdot \left (1-{M}\right)\label{1}\\
B=I_{in}\cdot {M}\label{2}
\end{gather}
to obtain the foreground car images \(X_{ref}\) and the ambient background images
\(B\). Then, we use the End-to-End Neural Renderer Plus (E2E-NRP) \(N\), our proposal neural rendering component,
to obtain the rendered 2D vehicle images \(X_{ren}\ \) through
\begin{gather}
X_{r e n}=N\left (Msh, UV_{adv}, \Phi_{c a m}, X_{r e f}\right)\text{,}\label{3}
\end{gather}
where \(Msh\) and \(UV_{adv}\) are the 3D mesh and the UV texture map of the
vehicle, respectively. To obtain the realistic vehicle pictures, we apply a simple transformation
\(I_{out} = X_{ren} + B\) to attach the foreground \(X_{ren}\) to the corresponding
background. We subsequently apply a Random Output Augmentation (ROA)\cite{Suryanto_2023_ICCV} to \(I_{out}\), which is designed to enhance texture robustness through random digital transformations, such as scaling, translation, brightness, and contrast. After that, we feed the output of ROA \(I_{ROA}\) into the target detector \(D\) and obtain the detection results represented as \(R = D(I_{ROA})\).

Our framework aims to generate adversarial camouflages for vehicles to evade the detection of the vehicle detector. We can obtain the final adversarial camouflage through the solution of a specific optimization problem denoted as
\begin{gather}
\underset{UV_{adv}} {\operatorname{argmax}} \mathcal{L}\left(D\left(O\left(N\left(Msh, UV_{adv}, \Phi_{cam}, X_{ref}\right)+B\right)\right), Y\right)\text{,}\label{4}
\end{gather}
where \(\mathcal{L}\) is our proposal loss function and \(O\) represents the ROA module.

\subsection{Multi-Weather Dataset}

According to Athalye et al.'s EOT study \cite{athalye2018synthesizing}, adding various weather conditions to training data notably boosts attack robustness. Nevertheless, real-world multi-weather dataset collection is hindered by high labor expenses, weather's inherent unpredictability, and difficulties in getting the vehicle's mask \(M\).

To address the above difficulties, we use CARLA \cite{dosovitskiy2017carla}, an autonomous driving simulation environment
based on Unreal Engine 4 (UE4) to obtain the multi-weather dataset. Modifying the weather and time parameters with CARLA API to simulate different weather and light environment conditions is convenient. Moreover, with its built-in semantic segmentation camera, we can accurately and conveniently segment foreground and background. 

\textbf{Base multi-weather dataset}: We have constructed a base multi-weather dataset for camouflage generation in our ICML paper. To simulate different conditions, we strategically vary the sun altitude angle and fog density. The sun altitude angle is instrumental in modulating the intensity of sunlight within the environment, influencing the light's interaction with the vehicle's camouflage. This variation in lighting can significantly affect the visibility and effectiveness of the camouflage. Concurrently, fog density, a critical environmental parameter, determines how much the vehicle's surface is obscured. At specific densities, fog can effectively render parts of the texture partially or entirely invisible, a factor that is crucial in determining the success of adversarial textures. Specifically, this dataset contains 16 different weather conditions created by combining four sun altitude angles (-90\degree, -30\degree, 30\degree, 90\degree) with four fog densities (0, 25, 50, 90). Within each weather
scenario, we randomly choose 20 car locations, and for each car location, imagery is captured from 128 different viewpoints. These viewpoints span every 45-degree increment in azimuth angle, four distinct altitude angles (0.0\degree, 22.5\degree, 45.0\degree, 67.5\degree), and four varying distances (5m, 10m, 15m, 20 m). In this dataset, we set the car surface material as a similar car paint material. Ultimately, we obtain a total of 40,960 images for texture generation.

\textbf{Enhanced multi-weather dataset}: We then optimize the multi-weather dataset in three aspects: (1) In Carla, the lighting conditions remain dark when the sun elevation angle is below 0\degree, so we adjust the sun elevation angle combination from (-90\degree, -30\degree, 30\degree, 90\degree)  in the base multi-weather dataset to (-90\degree, 10\degree, 45\degree, 90\degree) to simulate more weather conditions; (2) We add random camera viewpoint perturbations to the previously fixed 128 points to enhance the diversity of camera perspectives in the dataset; (3) We employ a material similar to our real-world camouflage implementation on the vehicle surface, specifically the sticker in place of the car paint. These three improvements can effectively enhance the robustness of the camouflage under various weather and viewpoint conditions. We finally obtain an enhanced multi-weather dataset comprising 32,768 high-quality images, which has fewer images but better quality than the base dataset. The comparison of the different datasets' effectiveness for camouflage generation is in Section \ref{datset-abl}. 

\subsection{End-to-End Neural Render Plus (E2E-NRP)}

In this section, we introduce E2E-NRP, a novel rendering
component that mitigates the limits of the previous neural-renderer-based camouflage methods. The world-align-based methods cannot accurately wrap the adversarial camouflage to the vehicle surface, which leads to suboptimal attack performance. Hence, we choose a UV-map-based neural renderer \cite{kato2018neural} as our base renderer to avoid this issue, as the optimized UV map can be accurately deployed on the vehicle.

However, this UV-map-based renderer has two problems. Firstly, it is inherently limited in its
ability to perform genuine end-to-end optimization of UV
maps, which necessitates post-processing and consequently leads to ineffective optimization. Secondly, it struggles to render the environmental characteristics on the vehicle surface, resulting in unrealistic rendered images. To address the first problem, we propose the End-to-End Neural Renderer, which can directly optimize the UV map. Regarding the second problem, we introduce the environment feature extractor (EFE) to achieve realistic environmental information rendering. Additionally, to reduce the retraining time of the EFE network for previously unseen vehicles, we train the EFE component to get a pre-trained EFE that can quickly adapt to unseen vehicles. In the following paragraphs, we explain the three components of our proposed E2E-NRP in detail: End-to-End Neural Renderer, Environment Feature Extractor, and pre-trained EFE.

\subsubsection{End-to-End Neural Renderer}

\begin{figure}[t]
\vfill
\begin{center}
\centerline{\includegraphics[width=\columnwidth]{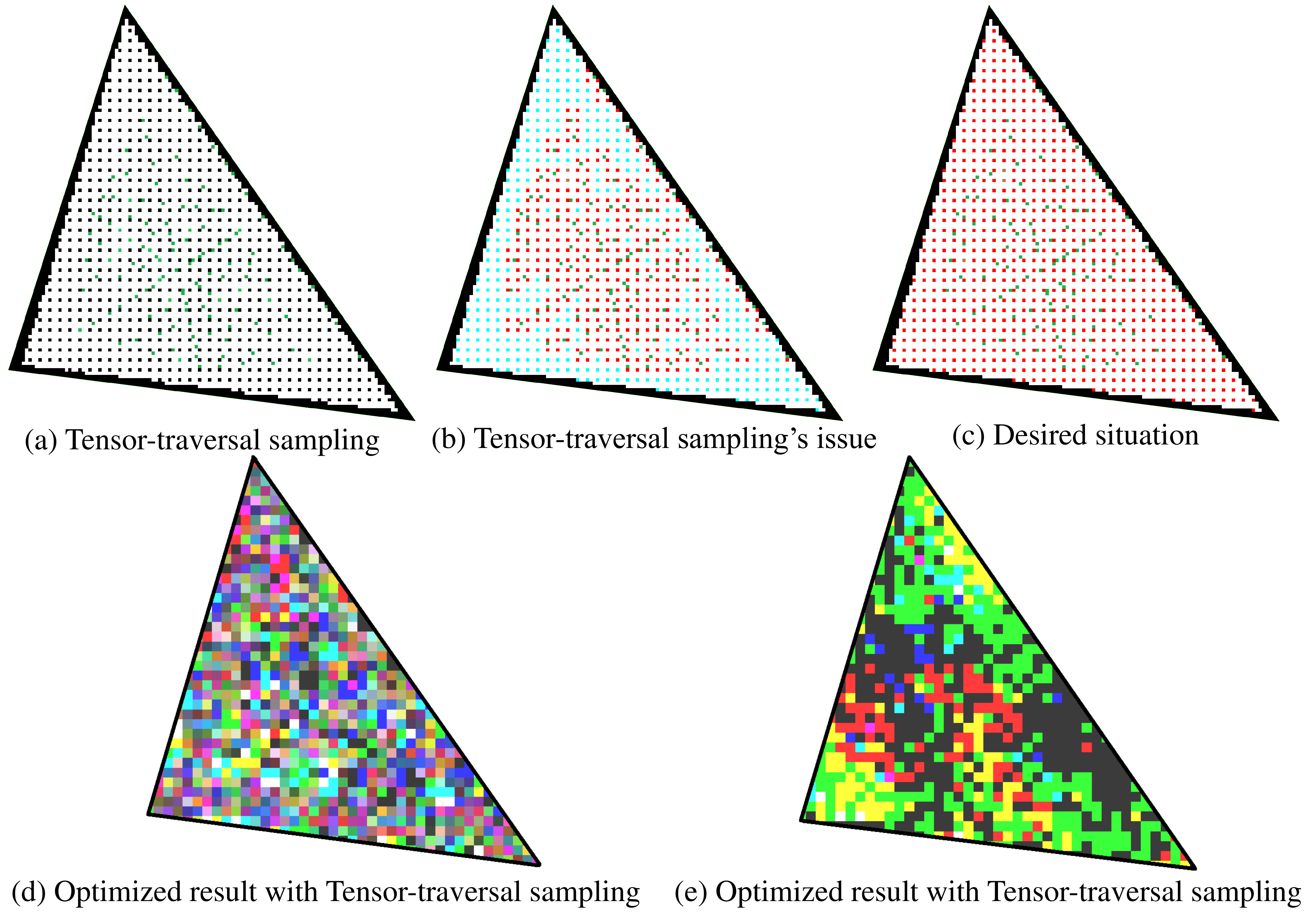}}
\caption{The illustration of tensor-traversal sampling and its issue. The triangle represents a facet's area in the UV map.
(a) illustrates the tensor-traversal sampling method's projection of a facet from \(T_{fc}\) onto the UV map. The projected positions of \(T_{fc}\) are represented by green points, while black points denote the pixel points on the UV map. (b) illustrates the issue of this sampling method: the red points are optimized during the camouflage generation, while the cyan points are not optimized. The numerous unoptimized points can lead to a decline in attack effectiveness. (c) is the desired situation in which all points on the UV map can be optimized. (d) shows the optimization result using tensor-traversal sampling, which leads to camouflage patterns resembling random noise with poor continuity. In contrast, (e) depicts the result of UV-travel sampling, exhibiting strong texture continuity and yielding more effective and robust attacks.}
\label{sampling}
\end{center}
\vfill
\end{figure}

Following \cite{wang2021dual,wang2022fca}, we use Neural Renderer (NR) \cite{kato2018neural} as our base renderer. The rendering process proceeds as follows: we feed 3D mesh \(Msh\), adversarial UV map  \(UV_{adv}\), and camera transformation \(\Phi_{cam}\) into this renderer. The input \(UV_{adv}\) is transformed to a facet-based texture tensor \(T_{fc}\), a face-aligned intermediate texture representation that maps the adversarial camouflage onto the 3D vehicle mesh. Subsequently, the camouflaged 3D mesh is rendered based on the camera transformation \(\Phi_{cam}\), producing an image of the camouflaged vehicle.

However, the original NR does not facilitate genuine end-to-end optimization of the UV map, as it only implements the forward propagation part of the sampling process from \(UV_{adv}\) to \(T_{fc}\) using CUDA code. As a result, NR alone cannot backpropagate gradients to the UV map. Therefore, the methods \cite{wang2021dual,wang2022fca} using NR optimize the facet-based texture tensor instead. This indirect optimization approach, however, compromises attack effectiveness because converting the optimized facet-based texture tensor into a deployable UV map involves non-differentiable post-processing steps, such as texture stretching and reshaping.
To address that, we first enable the gradient back-propagate to the UV map and then propose a new sampling method to achieve genuine end-to-end optimization.

\textbf{Gradient back-propogation to the UV map}: We develop the back-propagation CUDA code based on its forward propagation. The specific forward propagation algorithm \cite{kato2018neural} details are presented in Algorithm \ref{traditional sampling algorithm}. We call this sampling method tensor-traversal sampling. This method transforms the UV map to the facet-based texture tensor by projecting the coordinates of the facet-based texture tensor onto the UV map and assigning the facet-based texture tensor's value by interpolating neighboring points within the UV map. Specifically, the facet-based texture tensor \(T_{fc}\) has a shape of [\(nf\), \(ts\), \(ts\), \(ts\), 3], where \(nf\) denotes facet count and \(ts\) signifies texture size and the final dimension corresponds to RGB values. The UV map tensor has a shape of [\(wt\), \(ht\), 3], with \(wt\) representing the width and \(ht\) height; similarly, its last dimension pertains to RGB values as well. To assign the RGB value for the point (\(m\), \(x\), \(y\), \(z\)) within \(T_{fc}\), coordinates are initially projected onto the UV map as depicted in Figure \ref{sampling} (a). In this illustration, black points indicate pixel locations on the UV map, while green points represent projected positions of the points from a facet within \(T_{fc}\) onto the UV map. Subsequently, the RGB values of (\(m\), \(x\), \(y\), \(z\)) are assigned by interpolating those surrounding four points of the projection position on the UV map as outlined in Algorithm \ref{traditional sampling algorithm}. The back-propagation CUDA code corresponds directly to the forward propagation code. Specifically, during back-propagation, the gradient from each point in \(T_{fc}\) is propagated to the corresponding four neighboring points on the UV map, weighted by the same interpolation weights \(w\) used during the forward propagation.

\renewcommand{\algorithmicrequire}{\textbf{Input:}}  
\renewcommand{\algorithmicensure}{\textbf{Output:}} 
\begin{algorithm}[h]
  \caption{Tensor-traversal Sampling Algorithm} 
  \label{traditional sampling algorithm}
  \begin{algorithmic}[1]
    \REQUIRE
      UV map $\boldsymbol{UV}$[\(wt\), \(ht\), 3], where \(wt\) and \(ht\) are the width and height of UV map respectively; facet-based texture tensor $\boldsymbol{T_{fc}}$[\(nf\), \(ts\), \(ts\), \(ts\), 3], where \(nf\) and \(ts\) are facet count and texture size respectively; 
    \ENSURE
      Sampled facet-based texture tensor $\boldsymbol{T_{fc}}$;
    \STATE  $\boldsymbol{T_{fc}}=\boldsymbol{0}$ 
    \FOR{$m \leftarrow 0; m \textless nf; m++ $}
        \FOR{$x \leftarrow 0; x \textless ts; x++ $}
            \FOR{$y \leftarrow 0; y \textless ts; y++ $}
                \FOR{$z \leftarrow 0; z \textless ts; z++ $} 
                    \STATE Project $(m, x, y, z)$ from $\boldsymbol{T_{fc}}$ to $\boldsymbol{UV}$: $(a, b)$
                    \STATE // \begin{footnotesize}get the coordinates of neighbour points of $(a, b)$ in $\boldsymbol{UV}$\end{footnotesize}
                    \STATE $p_{0}= (\lfloor a \rfloor,\lfloor b \rfloor)$
                    \STATE $p_{1}= (\lfloor a \rfloor+1,\lfloor b \rfloor)$
                    \STATE $p_{2}= (\lfloor a \rfloor,\lfloor b \rfloor+1)$
                    \STATE $p_{3}= (\lfloor a \rfloor+1,\lfloor b \rfloor+1)$
                    \STATE // \begin{footnotesize}obatin the RGB value of $\boldsymbol{T_{fc}}[m, x, y, z]$\end{footnotesize}
                    \FOR {$i\leftarrow 0; i \textless 4; i++ $}
                    \STATE $w=\Vert p_{i}[0],a\Vert_1 *  \Vert p_{i}[1],b\Vert_1$
                    \STATE $\boldsymbol{T_{fc}}[m, x, y, z]+=w*\boldsymbol{UV}[p_{i}[0], p_{i}[1]]$
                    \ENDFOR  
                \ENDFOR
            \ENDFOR
        \ENDFOR
    \ENDFOR
  \end{algorithmic}
\end{algorithm}

\textbf{Sampling method enhancement}: However, the original sampling method employed in NR poses a significant challenge: due to the implementation of the sampling process through traversal of \(T_{fc}\), it can lead to a situation where no points from 
\(T_{fc}\) project onto the surrounding area of certain points on the UV map. As a result, these points on the UV map are neglected during sampling and remain unoptimized in the camouflage generation process. As shown in Figure \ref{sampling} (b), the green points are projections of the points from a facet within the facet-based texture tensor, while the red points indicate those points that can be optimized, and the cyan points denote those points that cannot be optimized. It shows that with the tensor-traversal sampling method, many points are not optimized, resulting in highly discontinuous optimization results as shown in Figure \ref{sampling} (d), which ultimately leads to suboptimal attack performance.

To address this problem, we enhance the previous tensor-traversal sampling process by traversing all points on the UV map to assign values for \(T_{fc}\), enforcing that every point on the UV map is optimized. We call this sampling method UV-traversal sampling. Specifically, for each point \((x, y)\) on the UV map, we first get its corresponding face index \(m\) in the facet-based texture tensor $T_{fc}$. Subsequently, we utilize the coordinates of the three vertices of that face in the UV map to compute the coordinates of that point in the facet-based texture tensor space, denoted as \((a, b, c)\). Following this step, we identify the eight immediately neighboring points based on the upper and lower boundaries of \(a, b,\) and \(c\). Furthermore, we distribute the RGB value of the point \((x, y)\)  on the UV map among these eight points according to the distance weighting scale \(w\). Finally, we perform normalization on all elements within the facet-based texture tensor. This approach ensures that all points on the UV map are incorporated into the tensor assignment, guaranteeing that each point can be optimized effectively. Detailed procedures are outlined in Algorithm \ref{novel sampling algorithm}. With this method, all points within the UV map can be optimized, as the desired situation illustrated in Figure \ref{sampling} (c).
The resulting optimization, shown in Figure \ref{sampling} (e), demonstrates significantly improved continuity, which contributes to more effective and robust camouflage generation.

\renewcommand{\algorithmicrequire}{\textbf{Input:}}  
\renewcommand{\algorithmicensure}{\textbf{Output:}} 
\begin{algorithm}[h]
  \caption{UV-traversal Sampling Algorithm} 
  \label{novel sampling algorithm}
  \begin{algorithmic}[1]
    \REQUIRE
      UV map $\boldsymbol{UV}$[\(wt\), \(ht\), 3], where \(wt\) and \(ht\) are the width and height of UV map respectively; facet-based texture tensor $\boldsymbol{T_{fc}}$[\(nf\), \(ts\), \(ts\), \(ts\), 3] and weight tensor $\boldsymbol{W$}[\(nf\), \(ts\), \(ts\), \(ts\)], where \(nf\) and \(ts\) are facet count and texture size respectively; 
    \ENSURE
      Sampled facet-based texture tensor $\boldsymbol{T_{fc}}$
    \STATE  $\boldsymbol{T_{fc}}=\boldsymbol{0},\boldsymbol{W}=\boldsymbol{0}$
                
    \FOR{$x \leftarrow 0; x \textless wt; x++ $}
        \FOR{$y \leftarrow 0; y \textless ht; y++ $}
             
                \STATE Project $(x, y)$ from $\boldsymbol{UV}$ to $\boldsymbol{T_{fc}}$: $(m, a, b, c)$
                \STATE // \begin{footnotesize}get the coordinates of neighbour points of $(a, b, c)$ in $\boldsymbol{T_{fc}}$\end{footnotesize}
                \STATE $p_{0}= (\lfloor a \rfloor,\lfloor b \rfloor,\lfloor c \rfloor)$
                \STATE $p_{1}= (\lfloor a \rfloor+1,\lfloor b \rfloor,\lfloor c \rfloor)$
                \STATE $p_{2}= (\lfloor a \rfloor,\lfloor b \rfloor+1,\lfloor c \rfloor)$
                \STATE $p_{3}= (\lfloor a \rfloor,\lfloor b \rfloor,\lfloor c \rfloor+1)$
                \STATE $p_{4}= (\lfloor a \rfloor+1,\lfloor b \rfloor+1,\lfloor c \rfloor)$
                \STATE $p_{5}= (\lfloor a \rfloor+1,\lfloor b \rfloor,\lfloor c \rfloor+1)$
                \STATE $p_{6}= (\lfloor a \rfloor,\lfloor b \rfloor+1,\lfloor c \rfloor+1)$
                \STATE $p_{7}= (\lfloor a \rfloor+1,\lfloor b \rfloor+1,\lfloor c \rfloor+1)$
                \STATE // \begin{footnotesize}distribute the RGB value of $\boldsymbol{UV}[x, y]$\end{footnotesize}
                \FOR {$i\leftarrow 0; i \textless 8; i++ $}
                    \STATE $w_{i}=\Vert p_{i}[0],a\Vert_1 *  \Vert p_{i}[1],b\Vert_1 * \Vert p_{i}[2],c\Vert_1$
                    \STATE $\boldsymbol{W}[m,p_{i}[0],p_{i}[1],p_{i}[2]]+= w_{i}$ 
                    \STATE $\boldsymbol{T_{fc}}[m, p_{i}[0], p_{i}[1], p_{i}[2]]+= w_{i}*\boldsymbol{UV}[x, y]$ 
                \ENDFOR  
        \ENDFOR
    \ENDFOR
    \STATE $\boldsymbol{W}$[\(nf\), \(ts\), \(ts\), \(ts\)] $\xrightarrow{\text{broadcast}}$ $\boldsymbol{W_{br}}$[\(nf\), \(ts\), \(ts\), \(ts\), 3]
    \STATE  $\boldsymbol{T_{fc}} = \boldsymbol{T_{fc}}\circ (1/\boldsymbol{W_{br}})$
                
  \end{algorithmic}
\end{algorithm}

\begin{figure}[t]
\vfill
\begin{center}
\centerline{\includegraphics[width=\columnwidth]{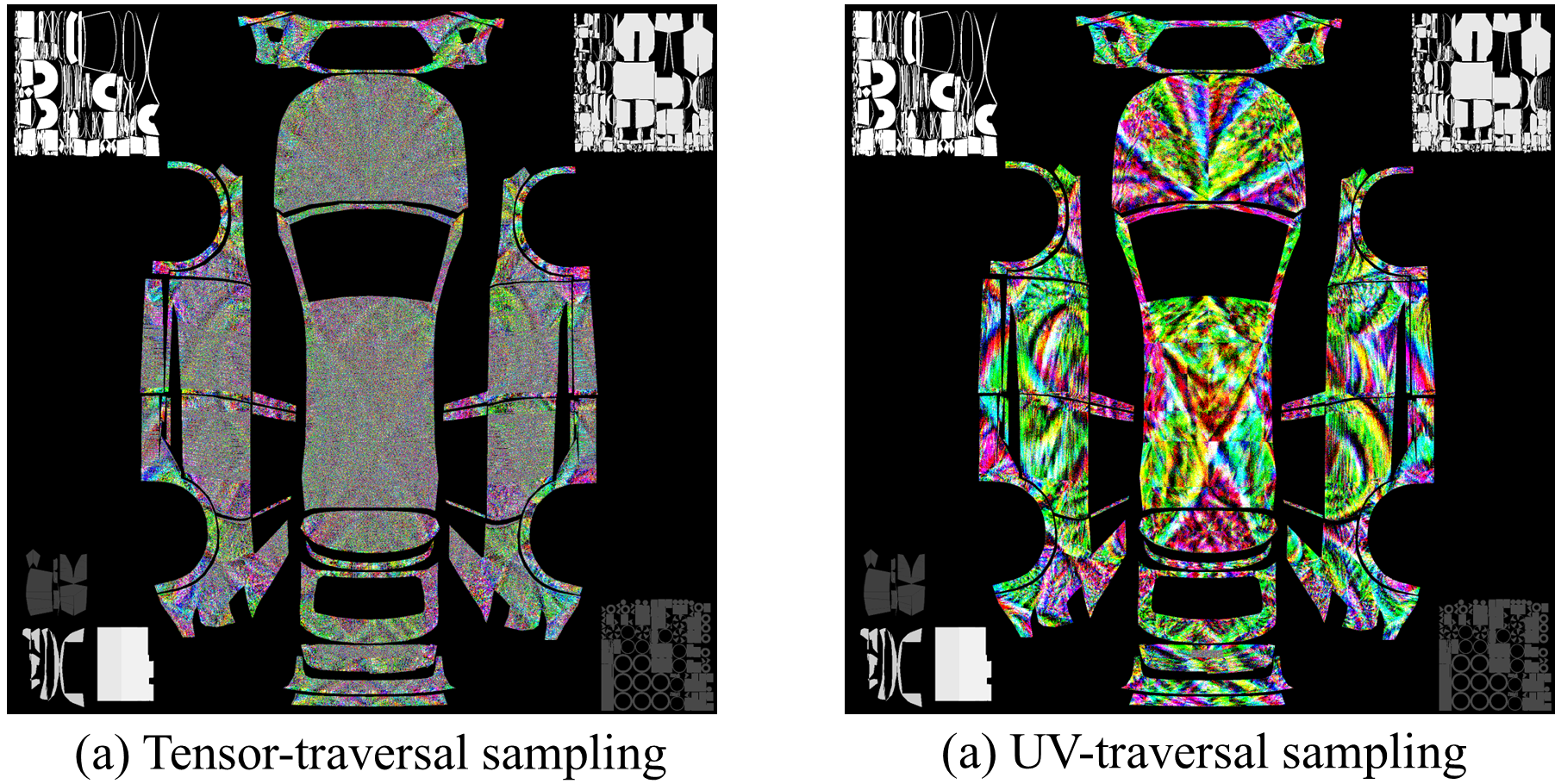}}
\caption{The comparison of adversarial camouflage generated by the neural renderer with different sampling methods. (a) Tensor-traversal sampling method: most areas appear as random noise. (b) Our proposed UV-traversal sampling method: most areas are strong adversarial texture patterns.}
\label{samplingResult}
\end{center}
\vfill
\end{figure}

Figure \ref{samplingResult} illustrates the differences in optimization results when NR employs these two sampling techniques. Figure \ref{samplingResult} (a) presents the camouflage obtained through the tensor-traversal sampling method, where the texture pattern exhibits poor quality and many areas of the UV map are random noise. This is because many points are not optimized and remain in their initial random noise states. In contrast, Figure \ref{samplingResult} (b) demonstrates that most areas of the UV map are effectively optimized with our proposed UV-traversal sampling approach. Visually, the UV map resembles a clear adversarial pattern rather than random noise, thanks to our proposed sampling method that effectively optimizes all the points.

\subsubsection{Environment Feature Exactor}

NR uses its own light source during rendering, which makes it impossible to render complex environmental characteristics similar to UE4. To amend this, we adopt the method proposed in
DTA and ACTIVE \cite{Suryanto_2022_CVPR,Suryanto_2023_ICCV}, introducing an encoder-decoder network to extract the
environmental characteristics in \(x_{ref}\). Since the NR's output
\(x_{nr}\) already encompasses the shape, rotation, and texture transformation of the vehicle, our network only needs to learn the
transformation of environmental characteristics. We call this network Environment Feature Xxtractor (EFE). EFE's outputs are two maps of environment
features EF. We can fuse them with \(x_{nr}\) through pixel-by-pixel
multiplication and addition to get \(x_{ren}\), an image of a textured
vehicle with environmental characteristics. 

\begin{figure}[t]
\vfill
\begin{center}
\centerline{\includegraphics[width=\columnwidth]{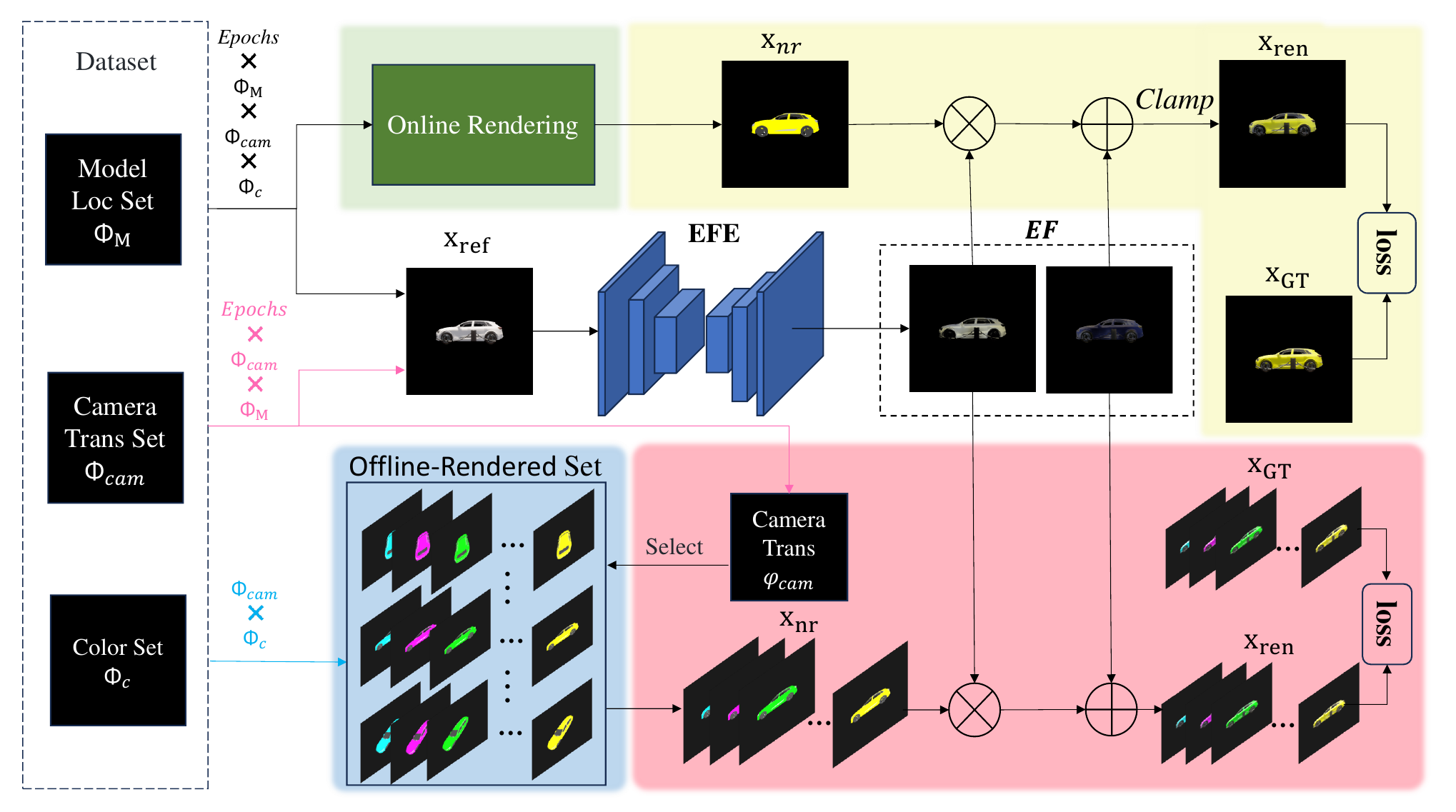}}
\caption{The comparison of EFE training methods in our base version and final version method. The blue and pink parts, respectively, represent enhancements to the green and yellow parts, which are part of the EFE training process in our base version method. In the original training process, the times of the EFE network inference and rendering both are \(num(Epochs)\)$\times$\(num(\Phi_{M})\)$\times$\(num(\Phi_{cam})\)$\times$\(num(\Phi_{c})\). We first transform the online rendering process (green part) into constructing an offline-rendered vehicle image set(blue part), thus reducing the rendering times to \(num(\Phi_{cam})\)$\times$\(num(\Phi_{c})\). Next, We use the output from a single EFE inference to combine with multiple color vehicle images (pink part), instead of using one output per color vehicle image as in the base version (yellow part). This reduces the number of EFE network inference times to \(num(Epochs)\)$\times$\(num(\Phi_{M})\)$\times$\(num(\Phi_{cam})\).}
\label{NRP_training_old}
\end{center}
\vfill
\end{figure}

\begin{table}[t]
\caption{Comparison of the EFE training time in our base and final version method. Values are one epoch training time.}
\label{EFE_training_time}
\begin{center}
\begin{small}
\begin{sc}
{
\begin{tabular}{lc}
\toprule
\multirow{1}*{\textbf{process}} & \multicolumn{1}{c}{\textbf{Time}}
     \\
\midrule
Base method                   & 120 min                  \\ 
Final method                  & 9 min   \\ 

\bottomrule
\end{tabular}
}
\end{sc}
\end{small}
\end{center}
\end{table}

Before using our framework to generate adversarial textures, we need to train EFE in advance. We first introduce the training process in our ICML version of the method, which is illustrated in the green and yellow part of Figure \ref{NRP_training_old}. The previous EFE training process involves the following inputs at each step: a masked image of the white car $x_{ref} \in X_{ref}$, fed into EFE to get a set of environment features \(EF\); a masked image of a car with one color $\varphi_{C}\in \Phi_{C}$ and the same vehicle location $\varphi_{M}\in \Phi_{M}$ and the same camera transformation $\varphi_{cam}\in \Phi_{cam}$ as $x_{ref}$, utilized to establish the ground truth \(x_{GT} \in X_{GT}\); a 3D mesh \(Msh\), the camera transformation $\varphi_{cam}$, and the color $\varphi_{C}$, performing online rendering to get the intermediate rendered result \(x_{nr}\). Subsequently, we fuse \(x_{nr}\) with \(EF\) to get the final rendered result \(x_{ren}\) and then compute loss between \(x_{ren}\) and \(x_{GT}\) to optimize the parameters of the EFE network. In this process, the times of the EFE network inference and rendering both are \(num(Epochs)\)$\times$\(num(\Phi_{M})\)$\times$\(num(\Phi_{cam})\)$\times$\(num(\Phi_{c})\), where \(num()\) denotes the counting function.

The training speed of the EFE is crucial to our framework. Therefore, we accelerate the training process of EFE in this version. The enhanced training process is illustrated in the blue and pink part of Figure \ref{NRP_training_old}. We introduce two strategies to enhance the training efficiency of EFE networks: the first mitigates the computations of the rendering process, and the second minimizes the computations of the EFE inference. Specifically, the first strategy entails the construction of an offline-rendered vehicle image set. Instead of rendering vehicle images online during EFE training, we render all of the vehicle images offline before EFE training to avoid re-rendering between different epochs. Meanwhile, the vehicle location \(\varphi_{M}\) within the simulation environment does not impact the output of the base neural renderer. Consequently, we perform a single rendering for various \(\varphi_{M}\) to eliminate redundant renderings of identical vehicle images. Eventually, the total times of renderings is reduced to \(num(\Phi_{cam})\)$\times$\(num(\Phi_{C})\). The second strategy uses the output from a single EFE inference to combine with multiple color vehicle images, instead of using one output per color vehicle image as in the ICML version. This approach reduces the number of EFE inferences required to achieve the same number of final rendering results. In summary, the times of EFE inference are reduced to \(num(Epochs)\)$\times$\(num(\Phi_{M})\)$\times$\(num(\Phi_{cam})\). To demonstrate our improvements in EFE training time, we separately train the EFE using the previous and enhanced processes, with a batch size of 16 on 54,728 images, and record the training time for each epoch. As shown in Table \ref{EFE_training_time}, our enhanced framework is 13 times faster compared to the previous one, demonstrating the effectiveness of the enhancements in reducing the EFE training time.

For the loss between the final rendered result \(x_{ren}\) and its corresponding ground truth \(x_{GT}\), we design the following function to optimize the EFE network:
\begin{gather}
L_{EFE} (x_{ref})=W\left (x_{r e f}\right) B C E\left (x_{ren}, x_{GT}\right),\label{5}
\end{gather}
where BCE is the binary cross-entropy loss, and \(W (x_{ref})=\frac{h*w}{s}\) is a weight function. \(h\) and \(w\) are the height and width of
\(x_{ref}\) and \(s\) is the number of pixel points in the vehicle part of
\(x_{ref}\). We introduce \(W (x_{ref})\) to balance NRP's rendering optimization across various camera viewpoints. The original BCE loss, calculated over the entire image, unfairly prioritizes rendering when the car occupies a small area of the entire image. To address this, we multiply BCE by \(W (x_{ref})\) to compute the difference over the vehicle area, improving rendering performance for views where the car occupies a small image area.

\begin{figure}[t]
\vfill
\begin{center}
\centerline{\includegraphics[width=\columnwidth]{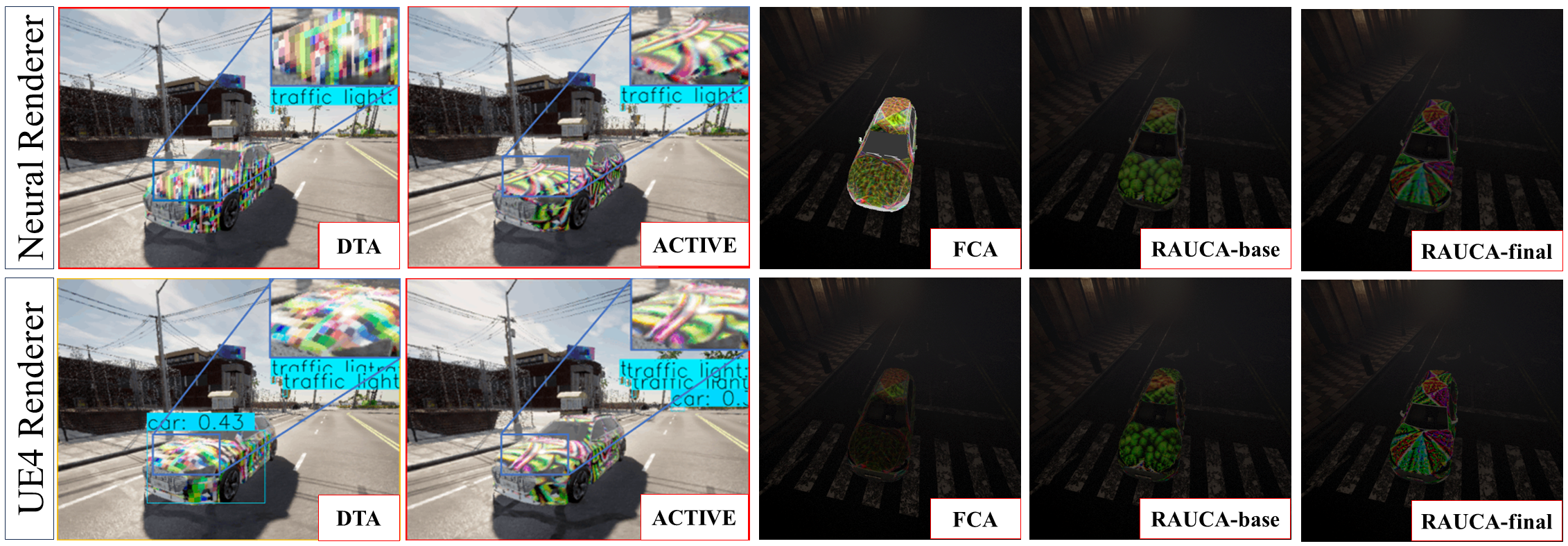}}
\caption{The comparison of rendering results of neural renderers used in different methods. The first row shows the results obtained by different neural renderers (already blended with the background), and the second row shows the rendered results in UE4. Our base and final renderers are the only implementations that execute both foreground environmental rendering and texture rendering similar to UE4.}
\label{render}
\end{center}
\vfill
\end{figure}

Compared to NR, the rendering component used in FCA \cite{wang2022fca} and DAS \cite{wang2021dual}, our E2E-NRP considers the
environmental characteristics to make the final rendered image more realistic. Furthermore, our renderer facilitates genuine end-to-end optimization. Compared to neural renderers used in DTA and ACTIVE \cite{Suryanto_2022_CVPR,Suryanto_2023_ICCV},
our rendering component can render the adversarial camouflage based on
UV mapping projection instead of world-aligned projection, with which the generated UV map can be accurately deployed, thus having better attack effectiveness. Besides renderers in other methods, we also compare the renderer proposed in our ICML paper, NRP. The only difference between NRP and our E2E-NRP is the neural renderer used to map textures to vehicles. The former uses the original NR, while the latter employs our improved E2E-NR. We show the rendering effects of different renderers within different attack methods in Figure \ref{render}. The renderings obtained by DTA and ACTIVE\textquotesingle s renderers have a noticeable difference in the vehicle textures from that in UE4. Additionally, the FCA\textquotesingle s renderer is not able to represent the complexity of light and shadow information, showing a clear difference between the foreground and the background. In contrast, the result of our rendering component E2E-NRP and NRP is relatively accurate both in terms of environmental characteristics and texture mapping.

\subsubsection{Pre-trained EFE}

\begin{figure}[t]
\vfill
\begin{center}
\centerline{\includegraphics[width=\columnwidth]{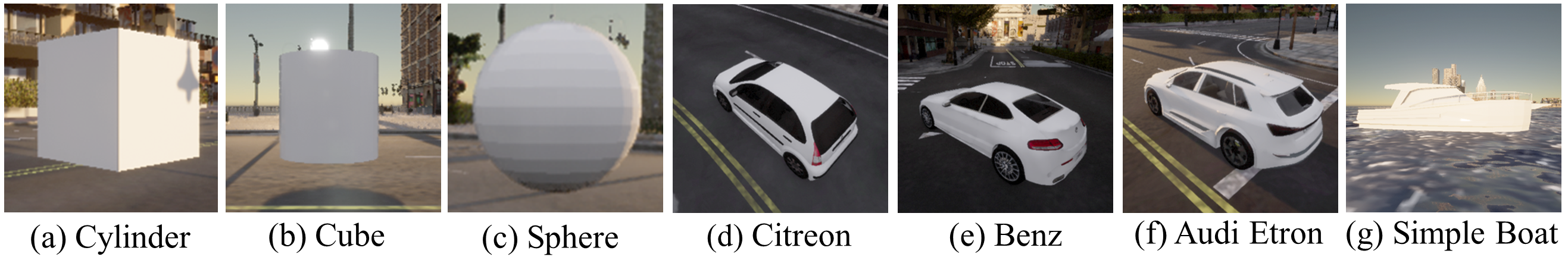}}
\caption{The different objects in the training dataset of the pre-trained EFE.}
\label{weight}
\end{center}
\vfill
\end{figure}

The EFE is the only component that needs to be retrained when we generalize our attack framework to other vehicles. To reduce the time cost of its training, We train the EFE with a dataset encompassing diverse objects to obtain the pre-trained EFE. Specifically, this dataset includes a cylinder, a cube, a square, three different cars (Benz, Citroen, and Audi E-tron), and a simple boat, shown in Figure \ref{weight}. The reason why we incorporate some basic objects (cylinder, cube, and square) into the dataset is to enhance the adaptability of the pre-trained EFE to unseen vehicle categories. With these objects, we generate 120,960 and 58,368 photo-realistic images for EFE training and testing. The details regarding these datasets can be found in Appendix B (in the supplementary materials). We train the EFE model for 20 epochs using the Adam optimizer with a learning rate of 0.01. We subsequently select the model exhibiting the best performance on the testing dataset as our pre-trained EFE. This pre-trained EFE can mitigate the training time for unseen vehicles. We validate this approach's effectiveness through related experiments in Section \ref{fine-tuning experiment}.

\subsection{Attack Loss}
We propose a novel attack loss function that consists of two key components for improving attack effectiveness. The first component is the Intersection Over Union (IOU) between the object detection model's output detection box and the ground-truth box; the second component comprises the class confidence score and objectiveness score of the output. The loss function is denoted as
\begin{gather}
H_d (x)=\operatorname{IoU}\left (H_b\left (x\right), g t\right) * H_c\left (x\right) * H_o\left (x\right)\notag \\
L_{a t k}\left (x\right)=-\log \left (1-\max \left (H_d (x)\right)\right),\label{6}
\end{gather}
where \(x\) is the input image of the target detector, \(H_{b} (x)\) is
the detection bounding box, \(gt\) is the
ground-truth box and \(IoU\left ( H_{b} (x),gt \right)\) is the Intersection
over Union (IoU) between \(H_{b} (x)\) and \(gt\). We use
\(IoU\left ( H_{b} (x),gt \right)\) as a weight term, which allows the
optimized loss function to focus more on the bounding box with a larger
intersecting ratio area with \(gt\). \(H_{o} (x)\) and \(H_{c} (x)\)
represent the objectiveness score and the car class confidence score for the
bounding box, respectively. \(H_{d}\left. (x \right.)\) is our
detection score, which is the product of the objectiveness confidence,
class confidence, and the intersect ratio. Due to 
 \(IoU\left ( H_{b} (x),gt \right)\), we only assign a non-zero detection score to the detected boxes which are intersected with the ground-truth box. This makes the texture optimization focus on making the target vehicle detection ineffective. We select the highest \(H_{d} (x)\) and use it to compute \(L_{atk} (x)\) through a log loss. By minimizing \(L_{atk} (x)\), we can make the camouflaged vehicle misclassified or undetected by the object detector. 

\subsection{Smooth Loss}
To ensure the smoothness of the generated texture for human
vision, we follow \cite{Crime} to utilize smooth loss 
\(L_{sm}\) to enhance texture consistency. Thanks to the end-to-end optimization, we can apply the smooth loss to the optimized UV map. The smooth loss function can be defined as:
\begin{gather}
L_{sm}=\frac{1}{N_{s m}} \sum_{i, j}\left (x_{i, j}-x_{i+1, j}\right)^2+\left (x_{i, j}-x_{i, j+1}\right)^2,\label{7}
\end{gather}
where \(x_{i,j}\) is the pixel value of \(x_{ren}\) at coordinate (i, j) and \(N_{s m}=H*W\) is a scale factor determined by the height, \(H\), and width, \(W\), of the UV map.

Finally, our loss function \(L_{total}\) can be summarized as
\begin{gather}
L_{total }=\alpha L_{a t k}+\beta L_{sm},\label{8}
\end{gather}
where $\alpha$\label{alpha}, $\beta$\label{beta}  are the hyperparameters to control the contribution of each
loss function.

\section{Experiments}
In this section, we first describe our experimental settings. Then we
evaluate the attack effectiveness
of our adversarial camouflage in both simulated
and physical environments.

\subsection{Experimental Settings}
\textbf{Datasets}: We utilize CARLA to generate datasets in our experimentation. To facilitate a comparative analysis with previous studies \cite{wang2021dual,wang2022fca, Suryanto_2022_CVPR, Suryanto_2023_ICCV}, we select the Audi E-Tron as the target vehicle model. We create datasets using various simulation settings. For Audi E-Tron's EFE training and testing, we utilize 73,728 and 65,536 photo-realistic images encompassing 16 distinct weather conditions, respectively. These conditions are a combination of four sun altitudes and four fog densities. Additionally, we generate 32,768 images for Audi E-tron's texture generation and 7,687 images for each adversarial camouflage used in robust evaluation. The weather conditions in these two datasets are identical to those used during the training and testing phases of the Audi E-Tron's EFE. For transferability assessment, we utilize an ``unseen-weather dataset" consisting of 7593 images in 16 novel weather conditions that are not seen during the texture generation phase. Additionally, we conduct real-world experiments by printing six types of adversarial camouflages and sticking them to scale models (1:12) of the Audi E-Tron car. We capture 960 photos under varying lighting conditions at different locations for each camouflage. Further details regarding the construction process of these datasets can be found in Appendix A, D, and F (in the supplementary materials).

\textbf{Baseline methods}: We compare our framework with the state-of-the-art adversarial camouflage methods: DAS \cite{wang2021dual}, FCA \cite{wang2022fca}, DTA \cite{Suryanto_2022_CVPR}, and ACTIVE \cite{Suryanto_2023_ICCV}. Both DAS and FCA are UV-map-based methods. They optimize the 3D texture by minimizing the attention-map-based scores and the detection scores of the detection model, respectively. Meanwhile, DTA and ACTIVE are both world-align-based methods. They optimize a square texture pattern with a neural network to project the texture onto the target vehicle.
In contrast, ACTIVE uses a projection network closer to the world-aligned projection in UE4. We compare our results using the official carefully optimized textures generated by these methods. Given the similarity in experimental setups between our approach and DAS/FCA and the generality of the textures generated by DTA and ACTIVE, we consider the comparison fair.

\textbf{RAUCA-base VS RAUCA-final}: We also compare the ICML version's method, RAUCA-base, with our current iteration, RAUCA-final. The differences between these two camouflage generation methods are as follows: Firstly, the base multi-weather dataset is utilized for texture generation in RAUCA-base, while the enhanced multi-weather dataset is employed in RAUCA-final. Secondly, RAUCA-base utilizes NRP as the renderer within its attack framework, while RAUCA-final employs E2E-NRP. Additionally, RAUCA-final includes a ROA module after rendering, which is absent in RAUCA-base. Moreover, the object used for computing smooth loss shifts from the rendering result (RAUCA-base) to the UV map (RAUCA-final).

\textbf{Evaluation metrics}\label{evaluation-metrics}: To evaluate the EFE rendering component, we use multiple metrics to quantify the difference between the final rendered image and the established ground truth, including Binary Cross-Entropy (BCE)\cite{Suryanto_2022_CVPR}, Mean Absolute Error (MAE), and Mean Squared Error (MSE). Furthermore, we evaluate the attack effectiveness of the adversarial camouflage with the AP@0.5 \cite{everingham2015pascal}, a standard benchmark reflecting both the recall and precision value when the detection IOU threshold is 0.5.

\textbf{Target detection models}\label{target-models}: To align with previous studies, we adopt YOLOv3 \cite{redmon2018YOLOv3} as the white-box target detection model for adversarial camouflage generation. To evaluate the effectiveness of the optimized camouflage, we utilize a suite of widely used object detection models treated as black-box models except for YOLOv3. This suite includes YOLOX \cite{ge2021yolox}, Deformable DETR (DDTR) \cite{zhu2020deformable}, Dynamic R-CNN (DRCN) \cite{zhang2020dynamic}, Sparse R-CNN (SRCN) \cite{sun2021sparse}, and Faster R-CNN (FrRCNN) \cite{faster-rcnn}, all of which are trained on the COCO dataset and implemented in MMDetection \cite{chen2019mmdetection}.

\textbf{Training details}\label{implementation-details}: We utilize the Adam optimizer with a learning rate of 0.01 for Audi E-tron's EFE training and texture generation. We train the EFE over a span of 40 epochs and select the model exhibiting the best performance on the testing dataset. We set the values of $\alpha$ and $\beta$ (refer to Eq. \ref{8}) at 1 and 0.01, respectively. For the mask of the vehicles, We directly obtain from the semantic segmentation camera feature in CARLA \cite{dosovitskiy2017carla}. During the adversarial camouflage generation phase, the camouflage texture is initialized randomly and trains with four epochs. We conduct experiments on a cluster with four NVIDIA RTX 3090 24GB GPUs.

\subsection{Evaluation in Physically-Based Simulation Settings}

\begin{table*}[t]
\caption{Performance comparison under seen-weather and unseen-weather settings. Values are car AP@0.5 (\%).}
\label{combined-table}
\begin{center}
\begin{small}
\begin{sc}
\resizebox{2\columnwidth}{!}{
\begin{tabular}{lccccccc|ccccccc}
\toprule
\multirow{3}*{\textbf{Methods}} 
& \multicolumn{7}{c|}{\textbf{Seen-weather}} 
& \multicolumn{7}{c}{\textbf{Unseen-weather}} 
\\

\cmidrule(lr){2-8} \cmidrule(lr){9-15}
& \multicolumn{3}{c}{\textbf{Single-stage}} & \multicolumn{3}{c}{\textbf{Two-stage}}& \multirow{2}*{Average} &
\multicolumn{3}{c}{\textbf{Single-stage}} & \multicolumn{3}{c}{\textbf{Two-stage}}& \multirow{2}*{Average}   \\

\cmidrule(lr){2-4}\cmidrule(lr){5-7}
\cmidrule(lr){9-11}\cmidrule(lr){12-14}
& YOLOv3 & YOLOX & DDTR & DRCN & SRCN & FrRCN & 
& YOLOv3 & YOLOX & DDTR & DRCN & SRCN & FrRCN & \\

\midrule
Normal         & 0.648 & 0.834 & 0.744 & 0.787 & 0.693 & 0.800 & 0.751 
               & 0.591 & 0.828 & 0.739 & 0.771 & 0.672 & 0.774 & 0.729 \\
Random         & 0.541 & 0.705 & 0.491 & 0.598 & 0.693 & 0.551 & 0.554 
               & 0.541 & 0.715 & 0.538 & 0.619 & 0.570 & 0.562 & 0.591 \\
DAS            & 0.605 & 0.790 & 0.659 & 0.675 & 0.654 & 0.717 & 0.683 
               & 0.545 & 0.786 & 0.682 & 0.659 & 0.642 & 0.710 & 0.671 \\
FCA            & 0.359 & 0.633 & 0.318 & 0.487 & 0.436 & 0.480 & 0.452 
               & 0.357 & 0.637 & 0.334 & 0.466 & 0.427 & 0.453 & 0.446 \\
DTA            & 0.317 & 0.449 & 0.139 & 0.419 & 0.244 & 0.291 & 0.310 
               & 0.346 & 0.470 & 0.163 & 0.454 & 0.280 & 0.291 & 0.287 \\
ACTIVE         & 0.173 & 0.338 & 0.129 & 0.268 & 0.190 & 0.219 & 0.220 
               & 0.176 & 0.372 & 0.130 & 0.289 & 0.213 & 0.201 & 0.230 \\
\midrule
RAUCA-base     & \underline{0.113} & \underline{0.318} & \underline{0.057} & \underline{0.261} & \underline{0.147} & \underline{0.202} & \underline{0.183} 
               & \underline{0.098} & \underline{0.292} & \underline{0.082} & \underline{0.249} & \underline{0.140} & \underline{0.166} & \underline{0.171} \\
RAUCA-final    & $\bm{0.010}$ & $\bm{0.134}$ & $\bm{0.029}$ & $\bm{0.159}$ & $\bm{0.076}$ & $\bm{0.085}$ & $\bm{0.082}$ 
               & $\bm{0.012}$ & $\bm{0.187}$ & $\bm{0.043}$ & $\bm{0.113}$ & $\bm{0.083}$ & $\bm{0.068}$ & $\bm{0.084}$ \\
\bottomrule
\end{tabular}
}
\end{sc}
\end{small}
\end{center}
\end{table*}

In this section, we conduct a comparative analysis of RAUCA against current advanced adversarial camouflage attack methods, including DAS, FCA, DTA, and ACTIVE.

\begin{figure*}[ht]
\begin{minipage}[t]{0.48\linewidth}
\centering
\includegraphics[width=\columnwidth]{{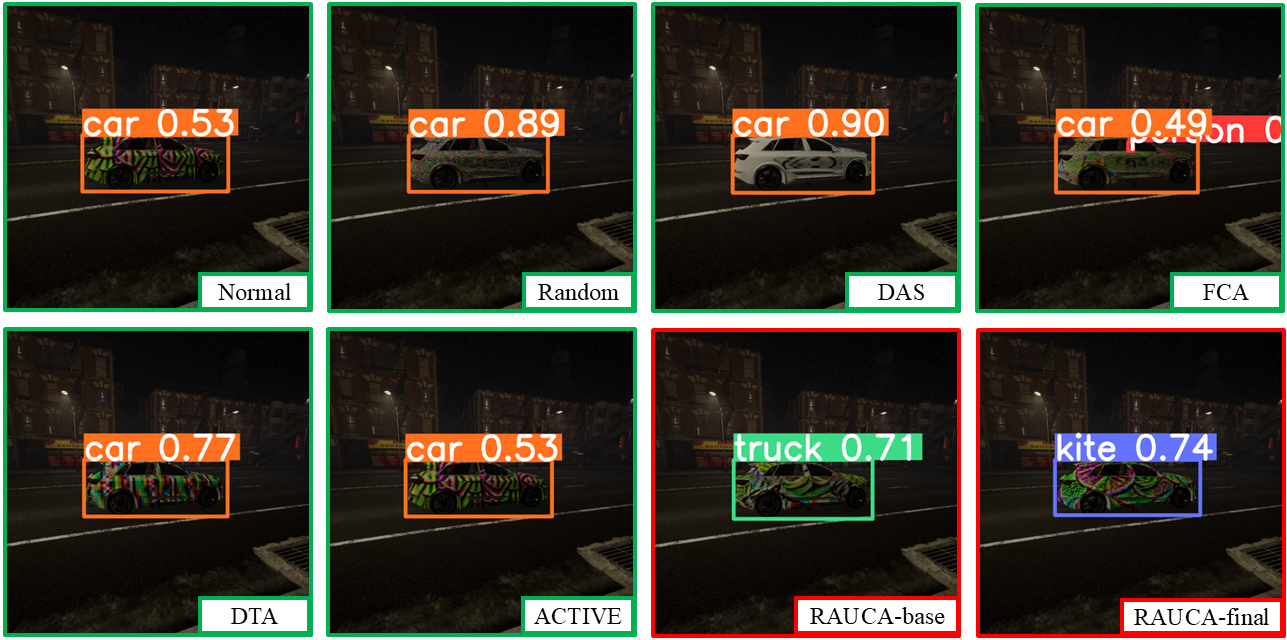}}

\caption{Attack comparison at night, a weather condition that has been included in our training set.}
\label{exp1}
\end{minipage}%
\hspace{0.6cm} 
\begin{minipage}[t]{0.48\linewidth}
\centering
\includegraphics[width=\columnwidth]{{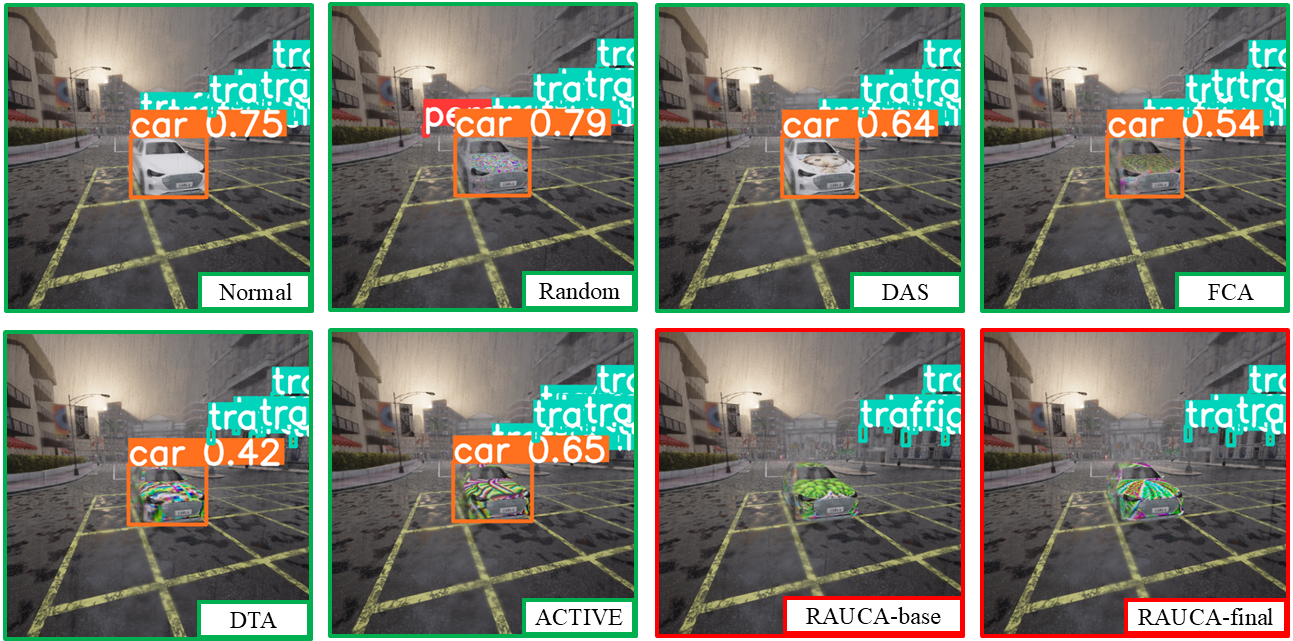}}

\caption{Attack comparison in the rainy day, a weather condition that hasn't appeared in our training set.}
\label{exp2}
\end{minipage}

\end{figure*}

\begin{figure}[t]
\vfill
\begin{center}
\centerline{\includegraphics[width=\columnwidth]{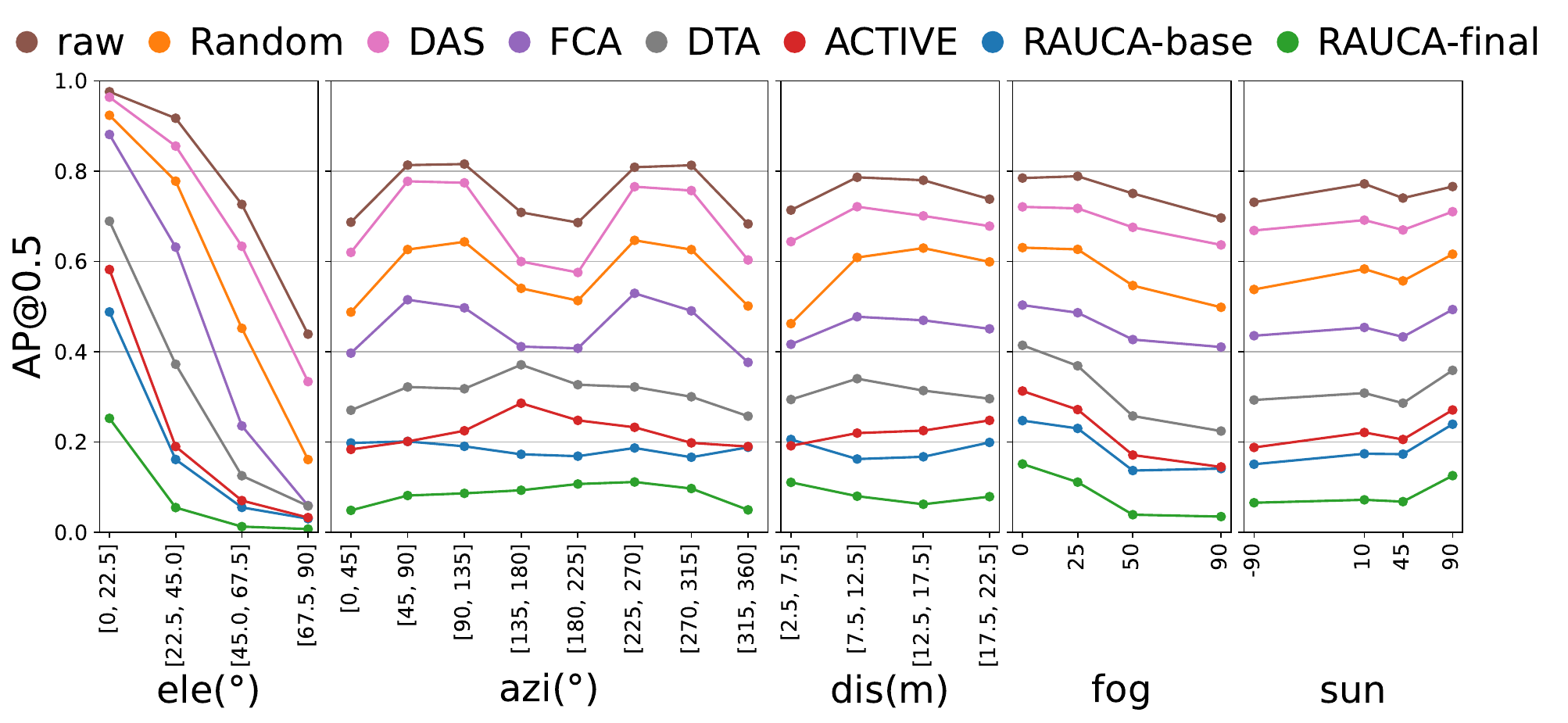}}

\caption{Attack comparison on different camera poses and weather parameters. ``ele" denotes elevation, ``azi" denotes azimuth, ``dis" denotes distance, ``fog" denotes fog density, and ``sun" denotes sun altitude angle. Values are car AP@0.5 (\%) averaged from all models.}
\label{curve}
\end{center}

\vfill
\end{figure}

\textbf{Robustness in multi-weather dataset}\label{4.2.1}: We run an extensive attack comparison using diverse detection models. In addition to YOLOv3, we use various black-box detection models to evaluate the camouflage.
The results are shown in the left part of Table \ref{combined-table}. We bold the most effective attack method and underline the second most effective approach. It shows that DAS performs even less effectively than random texture, primarily due to the limitations of partially painted camouflage. Meanwhile, FCA exhibits sub-optimal performance, only slightly better than random camouflage, because it cannot render sophisticated environment characteristics during texture generation. DTA and ACTIVE are also less effective than our methods because they do not consider multi-weather conditions during texture generation. Besides, their texture projection is not consistent between generation and testing. Our attack methods, RAUCA-base and RAUCA-final, outperform other baseline methods significantly. Our RAUCA-final, demonstrates a nearly 16\% improvement in attack effectiveness compared to other methods on the white-box YOLOv3 detector. Furthermore, it achieves an approximate 14\% improvement in the average score when considering other black-box detectors. Compared to RAUCA-base, the camouflage generated with RAUCA-final has a 10\% enhancement in attack performance, thanks to the improvements we have made to the attack framework and the multi-weather dataset. Figure \ref{exp1} shows an example of different vehicle camouflage at night: our RAUCA-final and RAUCA-base are still effective, while other methods fail to attack. 

Figure \ref{curve} shows the summarized performance of each camera transformation and
weather parameter; values are car AP@0.5 averaged from
the detectors used in Table \ref{combined-table}. We can see that RAUCA-final significantly outperforms other approaches on all of the camera transformations and weather conditions, while RAUCA-base performs better than previous methods in most cases. Our final method improves multi-view robustness over previous methods on all the viewpoints, thanks to the end-to-end UV mapping-based projection for our textures and effective rendering of the environmental characteristics in the foreground. In addition, due to incorporating a multi-weather dataset during camouflage generation, our method is also the most robust in different weather environments.

\hypertarget{4.2.2}{\textbf{Transferability in unseen-weather dataset}}\label{unseen-weather dataset}: We further evaluate the
transferability of our camouflage under unseen weather conditions.
As shown in the right part of Table \ref{combined-table}, our methods, RAUCA-base and RAUCA-final, achieve the best attack performance on all types of detectors. Our RAUCA-final outperforms the previous state-of-the-art method, ACTIVE, by 16\% on the white-box detector YOLOv3, and approximately in average of 15\% considered other black-box detectors, which highlights the transferability of our approach across diverse weather conditions. Such good transferability stems from the use of the multi-weather dataset with various fog levels and lighting conditions for texture optimization. The dataset includes 16 different weather conditions, including extreme conditions, such as dark light and dense fog. In the dense fog weather, the vehicle camouflage is heavily obscured but still effective against detectors. The weather in real life is typically not so extreme, so our camouflage can transfer well and achieve effective attacking performance in the unseen conditions. 

Figure \ref{exp2} shows an example of different camouflage on rainy days. Notice that rainy weather is excluded from texture generation. Even though the rain blurs part of the camouflage and the camera, our RAUCA-final and RAUCA-base succeed in attacking the target detector, whereas other methods fail to attack.

\begin{table*}[t]
\caption{Performance comparison under defense methods LGS and PAD. Values are car AP@0.5 (\%).}
\label{defense-table}
\begin{center}
\begin{small}
\begin{sc}
\resizebox{2\columnwidth}{!}{
\begin{tabular}{lccccccc|ccccccc}
\toprule
\multirow{3}*{\textbf{Methods}} 
& \multicolumn{7}{c|}{\textbf{LGS \cite{naseer2019local}}} 
& \multicolumn{7}{c}{\textbf{PAD \cite{jing2024pad}}} 
\\

\cmidrule(lr){2-8} \cmidrule(lr){9-15}
& \multicolumn{3}{c}{\textbf{Single-stage}} & \multicolumn{3}{c}{\textbf{Two-stage}}& \multirow{2}*{Average} &
\multicolumn{3}{c}{\textbf{Single-stage}} & \multicolumn{3}{c}{\textbf{Two-stage}}& \multirow{2}*{Average}   \\

\cmidrule(lr){2-4}\cmidrule(lr){5-7}
\cmidrule(lr){9-11}\cmidrule(lr){12-14}
& YOLOv3 & YOLOX & DDTR & DRCN & SRCN & FrRCN & 
& YOLOv3 & YOLOX & DDTR & DRCN & SRCN & FrRCN & \\

\midrule
Normal         &  0.643   & 0.840    & 0.741    & 0.773    & 0.677    & 0.791    & 0.744 & 0.598  & 0.797  & 0.701  & 0.746  & 0.648  & 0.756  & 0.708  \\
Random         & 0.532 & 0.700 & 0.472 & 0.592 & 0.533 & 0.539 & 0.561 & 0.311  & 0.414  & 0.298  & 0.373  & 0.280  & 0.319  & 0.333  \\
DAS            & 0.594 & 0.795 & 0.649 & 0.659 & 0.624 & 0.697 & 0.670 & 0.438  & 0.577  & 0.508  & 0.529  & 0.462  & 0.526  & 0.507  \\
FCA            & 0.350 & 0.628 & 0.310 & 0.482 & 0.421 & 0.471 & 0.444 & 0.214  & 0.392  & 0.215  & 0.328  & 0.230  & 0.301  & 0.280  \\
DTA            & 0.309 & 0.458 & 0.129 & 0.410 & 0.222 & 0.267 & 0.299 & 0.193  & 0.297  & 0.107  & 0.271  & 0.121  & 0.200  & 0.198  \\
ACTIVE         & 0.166 & 0.335 & 0.114 & 0.266 & 0.171 & 0.206 & 0.210 & 0.121  & 0.250  & 0.100  & \underline{0.204}  & 0.101  & 0.153  & 0.155 \\
\midrule
RAUCA-base     &\underline{0.095} & \underline{0.308} & \underline{0.056} & \underline{0.249} & \underline{0.125} & \underline{0.184} & \underline{0.170} & \underline{0.063} & \underline{0.227} & \underline{0.073} & 0.212 & \underline{0.085} & \underline{0.140} & \underline{0.133} \\
RAUCA-final    & $\bm{0.017}$ & $\bm{0.132}$ & $\bm{0.033}$ & $\bm{0.160}$ & $\bm{0.073}$ & $\bm{0.077}$ & $\bm{0.082}$ & $\bm{0.020}$ & $\bm{0.139}$ & $\bm{0.054}$ & $\bm{0.158}$ & $\bm{0.047}$ & $\bm{0.089}$ & $\bm{0.084}$ \\
\bottomrule
\end{tabular}
}
\end{sc}
\end{small}
\end{center}
\end{table*}

\textbf{Robustness against defense techniques}: In this section, we evaluate the robustness of our camouflage against defense methods. We adopt two established defense techniques, LGS \cite{naseer2019local} and PAD \cite{jing2024pad}. These defenses are applied to images from the seen-weather dataset and the processed images are then evaluated using different object detectors. As shown in Table \ref{defense-table}, RAUCA-Final consistently maintains strong attack performance across all detectors, with minimal degradation after applying defenses. This is because traditional defenses mainly target adversarial patches, while our full-body camouflage comprehensively covers the vehicle. As a result, LGS shows limited defensive capability, with our camouflage still achieving successful attacks, demonstrating its robustness even under this defense strategy. For PAD, two scenarios are observed: one where the camouflage is not detected and thus the defense fails, and another where the camouflage is detected and the defense succeeds. However, in the latter case, since the entire vehicle is masked out due to the extensive coverage of the camouflage, the object detectors still fail to accurately detect the target vehicle after the defense.

\subsection{Evaluation in Real-World Settings}

\begin{figure}[t]
\vfill
\begin{center}
\centerline{\includegraphics[width=\columnwidth]{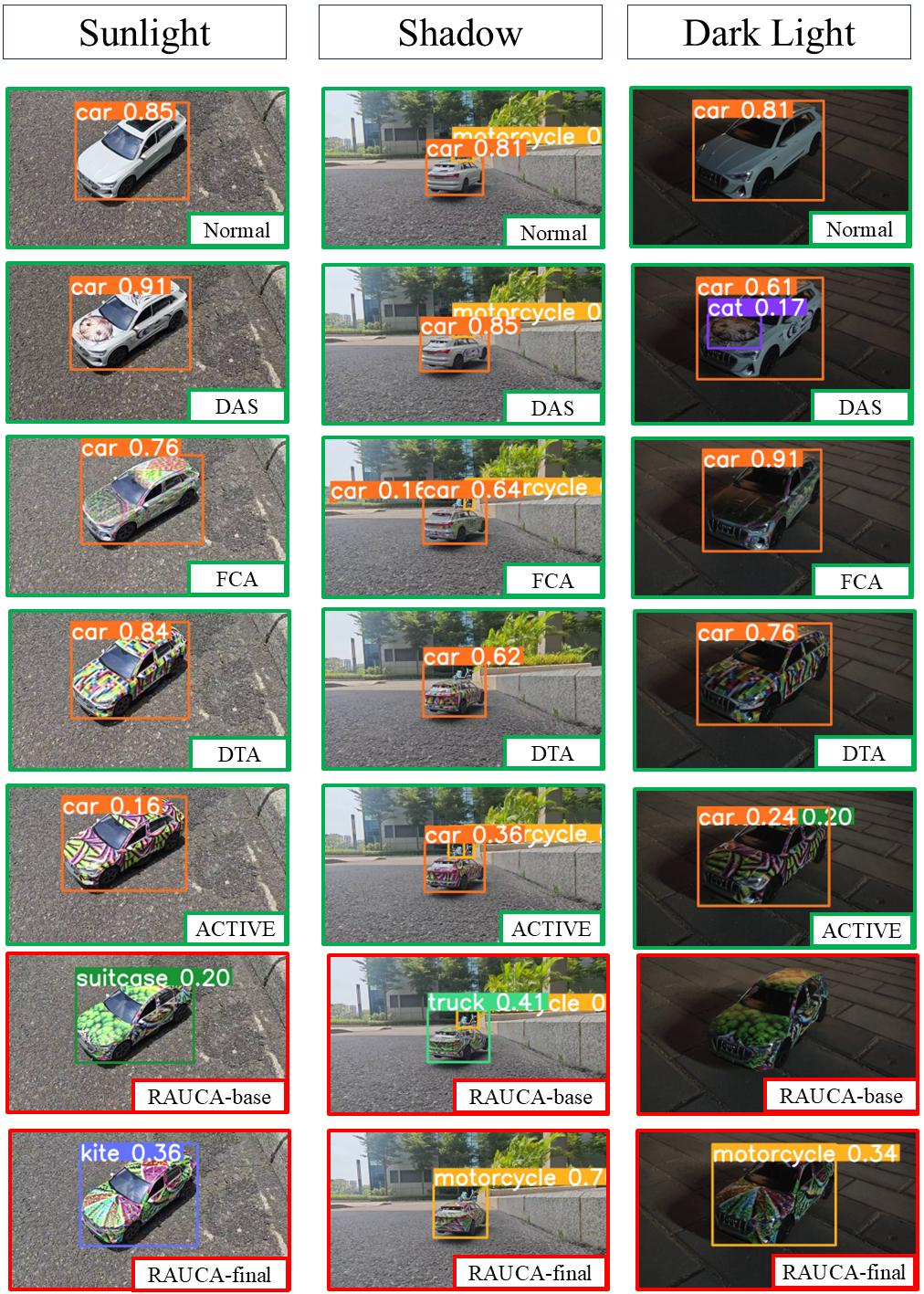}}
\caption{Real-world evaluation using six different types of camouflage in different light conditions.}
\label{real_world}
\end{center}
\vfill
\end{figure}

In this section, we move our test to the real world. We evaluate the attack performance of different adversarial camouflages in three light conditions: sunlight, shadow, and dark light, along with distinct solar altitude angles (near 90° and near 0°). We test all of the camouflages on different detectors, and the results are presented in Table \ref{real-table1}. We bold the most effective attack method and underline the second most effective approach. Our RAUCA-final achieves the highest attack effectiveness. Notably, RAUCA-final improves the averaged attack effectiveness by 3\% compared to the state-of-the-art method ACTIVE, showcasing our method's significant progress and potential in different environments.
Figure \ref{real_world} shows the attack examples under different light conditions. Our RAUCA-base and RAUCA-final consistently deceive the target detector across diverse lighting conditions in these examples, while other methods fail to attack.

\begin{table}[t]
\caption{Attack comparison in the real world. Values are the car AP@0.5 (\%)}.
\label{real-table1}
\begin{center}
\begin{small}
\begin{sc}
\resizebox{1\columnwidth}{!}{
\begin{tabular}{lccccccc}
\toprule
\multirow{2}*{\textbf{Methods}} & \multicolumn{3}{c}{\textbf{Single-stage}}& \multicolumn{3}{c}{\textbf{Two-stage}} & \multirow{2}*{Average}  \\ 
\cmidrule(lr){2-4}\cmidrule(lr){5-7}
                         & YOLOv3            & YOLOX                & DDTR                 & DRCN              & SRCN                & FrRCN          \\ 
\midrule
Normal       & 0.869            & 0.900            & 0.960            & 0.963            & 0.953            & 0.964            & 0.935 \\
DAS          & 0.862            & 0.909            & 0.932            & 0.939            & 0.940            & 0.916            & 0.916 \\
FCA          & 0.634            & 0.858            & 0.700            & 0.836            & 0.799            & 0.792            & 0.770 \\
DTA          & 0.519            & 0.777            & 0.561            & 0.740            & 0.740            & 0.667            & 0.667 \\
ACTIVE       & 0.359            & \underline{0.545} & $\bm{0.353}$     & \underline{0.535} & 0.586            & \underline{0.464} & \underline{0.474} \\ 
\midrule
RAUCA-base   & \underline{0.312} & 0.611            & 0.382            & 0.547            & \underline{0.581} & $\bm{0.458}$     & 0.482 \\
RAUCA-final  & $\bm{0.241}$     & $\bm{0.500}$     & \underline{0.355} & $\bm{0.486}$     & $\bm{0.580}$     & 0.484            & $\bm{0.441}$ \\
\bottomrule
\end{tabular}
}
\end{sc}
\end{small}
\end{center}
\end{table}

\subsection{Evaluation of the Pre-trained EFE}\label{fine-tuning experiment}

\begin{table}[t]
\caption{Comparison of the effect of using pre-trained EFE on training convergence epochs.}
\label{finetune_time}
\begin{center}
\begin{small}
\begin{sc}
\resizebox{1\columnwidth}{!}{
\begin{tabular}{lcccc}
\toprule
\multirow{1}*{\textbf{process}} 
 & AudiA2 & NissanPatrol & Sophisticated & Plane \\
\midrule
Without the pre-trained EFE                      & 26   & 51   & 44   & 132      \\ 
With the pre-trained EFE                    & 5     & 10  & 8   & 24         \\ 

\bottomrule
\end{tabular}
}
\end{sc}
\end{small}
\end{center}
\end{table}

\begin{figure}[ht]
\vfill
\begin{center}
\centerline{\includegraphics[width=\columnwidth]{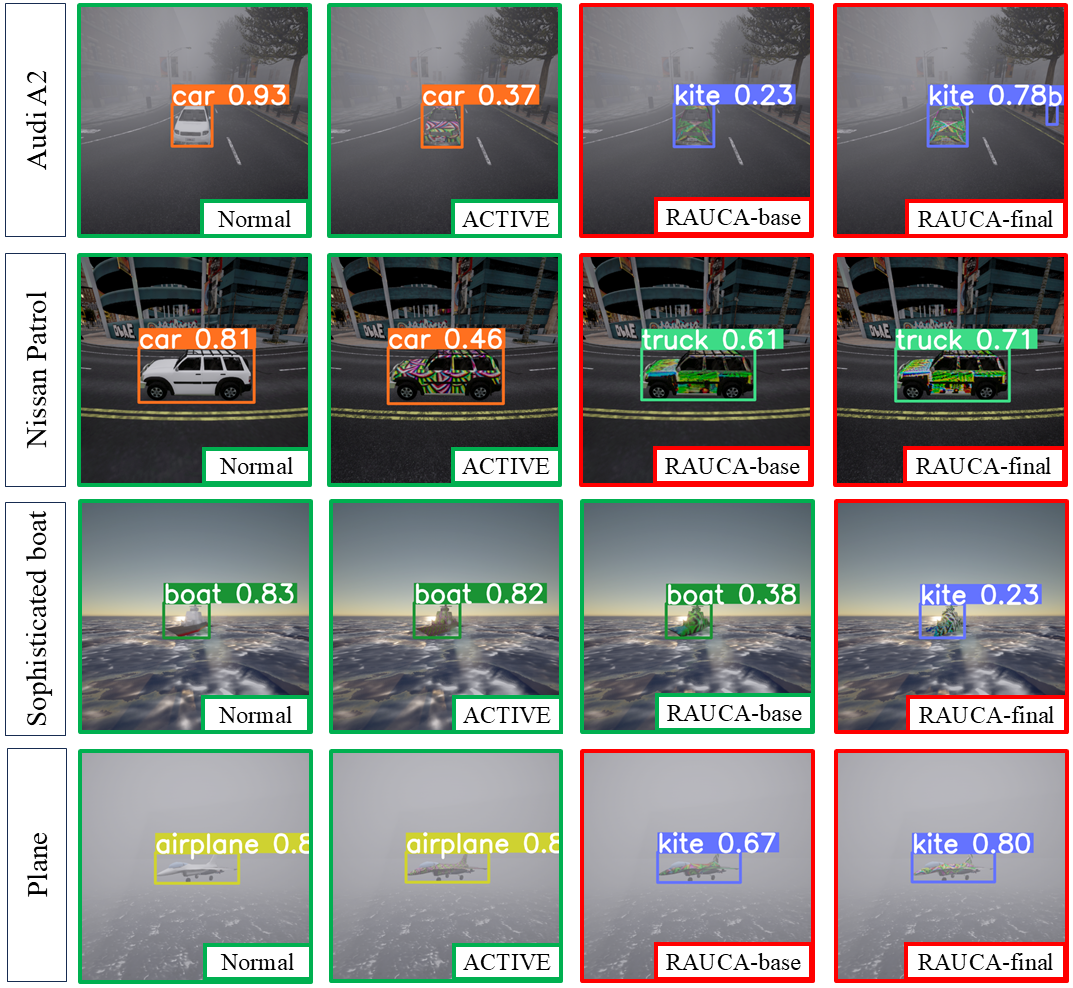}}
\caption{Attack comparison on four vehicles, i.e. Audi A2, Nissan Patrol, a sophisticated boat, and a plane.}
\label{all_object}
\end{center}
\vfill
\end{figure}

In this section, we evaluate the impact of our pre-trained EFE. We compare the convergence speed of EFE training on unseen vehicles, starting from the pre-trained EFE versus starting from scratch. Our experiment encompasses four vehicles: Audi A2, Nissan Patrol, a sophisticated boat, and a plane. The four vehicles exhibit different characteristics: Audi A2 bears similarity to the car named Citeron within the pre-training dataset for the EFE; Nissan Patrol differs from all of the cars present in the pre-training dataset; the sophisticated boat represents a boat model which has a more complex structure compared with the simple boat model in the training dataset; the plane introduces an entirely novel category of vehicles with no analogous items in the pre-training dataset. For each target vehicle, we produce 9,504 images for training and 233,472 images for testing. The details regarding these datasets can be found in Appendix C (in the supplementary materials). 

We analyze the number of epochs required for training to converge. The criteria for convergence are as follows: When training reaches epoch \(i\), if the difference between the loss over the previous five epochs and that at epoch \(i\) is within a hyperparameter $\eta$ (in our case, set to 0.003), then epoch \(i\) is considered a converged point. The loss computation process is carried out as follows: First, we train the EFE using the Adam optimizer with a learning rate of 0.01. For each epoch's checkpoint, we evaluate its performance on the test dataset, calculating the Mean Absolute Error (MAE) loss \(S\) between the rendered image and the corresponding ground truth. For epoch \(i\) , we use $MS_{i}=\gamma*S_{i-1} + (1 - \gamma)*S_{i}$ to smooth the loss, where $\gamma$ is the smoothing parameter, set to 0.5. We use the smoothed loss \(MS_{i}\)  to determine whether the training has converged. The experimental results are presented in Table \ref{finetune_time}, demonstrating that employing the pre-trained EFE effectively reduces the number of epochs required for convergence across all scenarios.

\begin{table}[t]
\caption{Attack comparison on different objects. Values are the respective class AP@0.5 (\%) averaged from six detectors.}
\label{sim-table4}
\begin{center}
\begin{small}
\begin{sc}
\resizebox{1\columnwidth}{!}{
\begin{tabular}{lccccccc}
\toprule
\multirow{1}*{\textbf{Methods}}     
                         & Audi A2            & Nissan Patrol      & Sophisticated boat      & Plane                    \\ 
\midrule
Normal                   & 0.701           & 0.622        & 0.858      &  0.949                   \\ 
ACTIVE                      & 0.303             & 0.404          & 0.698
      & 0.288                      \\ \midrule
RAUCA-base                   & 0.150             & 0.224         & 0.330
      & 0.266                    \\ 
RAUCA-final                     & $\bm{0.111}$              & $\bm{0.188}$    & $\bm{0.145}$              & $\bm{0.155}$                    \\ 
\bottomrule
\end{tabular}
}
\end{sc}
\end{small}
\end{center}
\end{table}

We then
 evaluate the effectiveness of the pre-trained EFE when it adapts to the above vehicles by finetuning in physically-based simulation settings. Utilizing the finetuned EFE, we apply our base and final attack framework on various vehicles to generate multiple camouflage. We also apply the state-of-the-art method ACTIVE\cite{Suryanto_2023_ICCV} on these vehicles for comparison. The details regarding the texture generation datasets and the evaluation datasets can be found in Appendix E (in the supplementary materials). The comparative results of the attack efficacy are presented in Table \ref{sim-table4}, with scores reflecting their respective AP@0.5 averaged from the six object detectors (See  \textbf{Target detection models} in Section \ref{target-models}). It shows that with the finetuned EFE, RAUCA-base and RAUCA-final can achieve strong attack performance on various vehicles, both better than ACTIVE. Figure \ref{all_object} illustrates examples of attacks on different objects. Our RAUCA-base and RAUCA-final frameworks generate camouflage that can successfully attack the detector on Audi A2, Nissan Patrol, and the plane targets. RAUCA-final's camouflages have better attack performance (higher confidence for wrong class). Notably, although only the RAUCA-final's camouflage successfully attacks the detector for the sophisticated boat, the camouflage generated by the RAUCA-base still effectively lowers the confidence of the correct class. The result demonstrates the effectiveness of the finetuned EFE.



\subsection{Ablation Studies}




\textbf{Effectiveness of different neural renderers for texture generation}: In this section, we perform an ablation experiment on various versions of renderers, including NR (FCA and DAS), NRP (RAUCA-base), and E2E-NRP (RAUCA-final). Notably, since there is no UV map involved in the optimization process for non-end-to-end renderers (NR and NRP), applying smooth loss to the UV map is not feasible; instead, we apply smooth loss with the weight parameter $\beta=0.01$ to the rendered image while adjusting the image height \(H\) and width \(W\) accordingly for NR and NRP.

The experiment results are presented in Table \ref{abl-renderer}. Our proposed NRP and E2E-NRP both demonstrate superior attack efficacy over the previous leading UV-map-based renderer, NR. Our E2E-NRP enhances average performance across all detection models by nearly 12\% compared to NR. Compared with NRP, our E2E-NRP demonstrates improvements in both the white-box model YOLOv3 and averaged performance considered other black-box models. Compared with NR, NRP incorporates EFE to render realistic environmental information, ensuring consistency between the foreground and background, thereby notably enhancing the effectiveness of the attack. Our proposed E2E-NRP further enhances the renderer's capability to backpropagate gradients to the UV map and optimizes the sampling method. This leads to an effective end-to-end optimization of all points on the UV map, thus achieving the best attack effectiveness among these renderers. 

It is important to emphasize that the benefit of end-to-end optimization is not just the enhancement of the attack effectiveness. Additionally, it allows incorporating some graphical constraints on UV map during the optimization process, such as naturalness, which may open new avenues for future advancements within the physical adversarial camouflage study.


\begin{table}[t]
\caption{The effectiveness of different neural renderers for texture generation. Values are the car AP@0.5 (\%).}
\label{abl-renderer}
\begin{center}
\begin{small}
\begin{sc}
\resizebox{1\columnwidth}{!}{
\begin{tabular}{lcccccccc}
\toprule
\multirow{2}*{\textbf{Methods}} &\multirow{2}*{\textbf{End-to-end}}& \multicolumn{3}{c}{\textbf{Single-stage}}& \multicolumn{3}{c}{\textbf{Two-stage}} & \multirow{2}*{Average}  \\ 
\cmidrule(lr){3-5}\cmidrule(lr){6-8}
               &   & YOLOv3            & YOLOX      & DDTR      & DRCN          & SRCN                     & FrRCN          \\ 
\midrule
NR-direct          &No   & 0.078              & 0.342         & 0.095      &  0.263           & 0.176               & 0.250      & 0.201        \\ 
 \midrule
NRP         &No  & 0.014              & 0.176          & $\bm{ 0.027}$      &  $\bm{ 0.151}$         & 0.077               & 0.100        & 0.091     \\


E2E-NRP      &Yes          & $\bm{0.010}$              & $\bm{0.134}$    & 0.029              & 0.159           & $\bm{0.076}$    & $\bm{0.085}$  & $\bm{0.082}$           \\ 
\bottomrule
\end{tabular}
}
\end{sc}
\end{small}
\end{center}
\end{table}

\begin{table}[t]
\caption{The effectiveness of different weather datasets for texture generation. Values are the car AP@0.5 (\%).}
\label{abl-dataset}
\begin{center}
\begin{small}
\begin{sc}
\resizebox{1\columnwidth}{!}{
\begin{tabular}{lccccccc}
\toprule
\multirow{2}*{\textbf{Methods}} & \multicolumn{3}{c}{\textbf{Single-stage}}& \multicolumn{3}{c}{\textbf{Two-stage}} & \multirow{2}*{Average}  \\ 
\cmidrule(lr){2-4}\cmidrule(lr){5-7}
                         & YOLOv3            & YOLOX      & DDTR      & DRCN          & SRCN                     & FrRCN          \\ 
\midrule
Single-weather Dataset                   & 0.108              & 0.422         & 0.063      & 0.348            & 0.180                & 0.279      & 0.233        \\ 
Base multi-weather Dataset                  & 0.063              & 0.901          & 0.054     &   0.311           & 0.093               & 0.195        & 0.17     \\ 
Enhanced multi-weather Dataset                                & $\bm{0.010}$              & $\bm{0.134}$    & $\bm{0.029}$              & $\bm{0.159}$           & $\bm{0.076}$   & $\bm{0.085}$  & $\bm{0.082}$          \\ 

\bottomrule
\end{tabular}
}
\end{sc}
\end{small}
\end{center}
\end{table}
\textbf{Effectiveness of different datasets for adversarial camouflage generation}\label{datset-abl}: 
In this section, we compare the effectiveness of different datasets for adversarial camouflage, including the single-weather dataset, the base multi-weather dataset, and the enhanced multi-weather dataset. The single-weather dataset has the same construction parameters as the base multi-weather dataset, except that the weather parameter is fixed at the default value in Carla. Compared with the base multi-weather dataset, our enhanced multi-weather dataset incorporates more weather conditions, utilizes a broader range of camera perspectives, and modifies the vehicle paint material to align with our physical camouflage implementation material. We use the above three datasets to generate camouflage and evaluate its attack effectiveness. The experimental results are shown in Table \ref{abl-dataset}, where we can see that the camouflage generated using our enhanced dataset outperforms those generated using the single-weather dataset and the base multi-weather dataset (an average of approximately 9\% across all detectors), demonstrating the effectiveness of our enhanced multi-weather dataset.

\begin{table}[t]
\caption{The effectiveness of the smooth loss for texture generation. Values
are the car AP@0.5 (\%).}
\label{abl-table3}
\begin{center}
\begin{small}
\begin{sc}
\resizebox{1\columnwidth}{!}{
\begin{tabular}{lccccccc}
\toprule
\multirow{2}*{\textbf{Methods}} & \multicolumn{3}{c}{\textbf{Single-stage}} & \multicolumn{3}{c}{\textbf{Two-stage}}  & \multirow{2}*{\textbf{Average}} \\ 
\cmidrule(lr){2-4}\cmidrule(lr){5-7}
                         & YOLOv3            & YOLOX     & DDTR       & DRCN          & SRCN           & FrRCN                   \\ 
\midrule
ACTIVE                      & 0.176              & 0.372          & 0.130      & 0.289             & 0.213                 & 0.201 & 0.230              \\
\midrule

$\beta$=100 & 0.019 & 0.216 & 0.045 & 0.174 & 0.076 & 0.126 & 0.109 \\ 
$\beta$=10 & 0.013 & 0.163 & 0.033 & 0.177 & 0.071 & 0.096 & 0.092 \\ 
$\beta$=1 & 0.011 & 0.164 & 0.038 & 0.183 & 0.063 & 0.091 & 0.092 \\ 
$\beta$=0.1 & 0.011 & 0.139 & 0.030 & 0.162 & 0.059 & 0.079 & 0.080 \\ 
$\beta$=0.01 & 0.010 & 0.134 & 0.029 & 0.159 & 0.076 & 0.085 & 0.082 \\ 
$\beta$=0.001 & 0.011 & 0.136 & 0.028 & 0.158 & 0.068 & 0.093 & 0.082 \\ 
$\beta$=0 & 0.038 & 0.298 & 0.073 & 0.230 & 0.130 & 0.186 & 0.160\\

\bottomrule
\end{tabular}

}
\end{sc}
\end{small}
\end{center}
\end{table}
\textbf{Effectiveness of the smooth loss for texture generation}: We examine the influence of smooth loss on the attack effectiveness of generated adversarial camouflage. We generate multiple camouflages by varying the hyperparameter $\beta$, where a larger $\beta$ increases the weight of the smooth loss in the total loss. The results are shown in Table \ref{abl-table3}, where values represent the car AP@0.5. It shows that the presence of the smooth loss is important, as setting $\beta$ to 0 results in a significant reduction in attack effectiveness. Furthermore, the attack efficacy of the generated camouflage remains robust across a broad range of $\beta$ values, consistently outperforming the state-of-the-art method, ACTIVE. However, excessively large $\beta$ can also lead to a certain decrease in effectiveness, as it causes the adversarial loss to constitute a smaller proportion of the total loss.

\begin{table}[t]
\caption{Comparing the impact \(W (x_{ref})\) on EFE training. Values are the correspond loss values between the final rendered result and ground truth.}
\label{abl-table0}
\begin{center}
\begin{small}
\begin{sc}
{
\begin{tabular}{lccccc}
\toprule
\multirow{1}*{\textbf{Methods}}  
& BCE & MAE & MSE       \\

\midrule
Without \(W (x_{ref})\)                   &  0.01864             & 0.00021                   & 0.00180               \\ 
With \(W (x_{ref})\)                  & $\bm{0.01847}$               & $\bm{0.00019}$                    & $\bm{0.00174}$                 \\ 

\bottomrule
\end{tabular}
}
\end{sc}
\end{small}
\end{center}
\end{table}

\textbf{Effectiveness of the \(W (x_{ref})\) for EFE training}:  During the training of the EFE, we utilize \(W (x_{ref})\), a weight function that deflates the loss according to the vehicle area in the image, to balance NRP's rendering optimization across various camera viewpoints. In this section, we evaluate the impact of \(W (X_{ref})\) on the accuracy of NRP rendering. We perform an ablation study to evaluate the test loss of the EFE trained with \(W(X_{ref})\) in comparison to those trained without this weight. We present the BCE, MSE, and MAE losses between the final rendered output and the ground truth. As illustrated in Table \ref{abl-table0}, incorporating \(W(X_{ref})\)  enhances rendering performance across all evaluated loss metrics, demonstrating the contribution of \(W(X_{ref})\) to the effectiveness of EFE.

\textbf{Effectiveness of the ROA component for texture generation}: 
In this section, we evaluate the impact of the ROA component integrated into our camouflage generation framework on the attack performance of the generated camouflage. As presented in Table \ref{ROA-abl}, the results demonstrate that incorporating the ROA component improves our camouflage attack by approximately 1\%. This indicates that the ROA component can boost the robustness of the generated camouflage, thereby highlighting ROA's effectiveness.

\begin{table}[t]
\caption{The effectiveness of the ROA component for texture generation. Values are the car AP@0.5 (\%).}
\label{ROA-abl}
\begin{center}
\begin{small}
\begin{sc}
\resizebox{1\columnwidth}{!}{
\begin{tabular}{lccccccc}
\toprule
\multirow{2}*{\textbf{Methods}} & \multicolumn{3}{c}{\textbf{Single-stage}}& \multicolumn{3}{c}{\textbf{Two-stage}} & \multirow{2}*{Average}  \\ 
\cmidrule(lr){2-4}\cmidrule(lr){5-7}
                         & YOLOv3            & YOLOX      & DDTR      & DRCN          & SRCN                     & FrRCN          \\ 
\midrule
Without ROA                   & 0.011              & 0.159         &    0.035   &   0.176          & 0.086                & 0.101      & 0.095        \\ 
WithROA                                & $\bm{0.010}$              & $\bm{0.134}$    & $\bm{0.029}$              & $\bm{0.159}$           & $\bm{0.076}$   & $\bm{0.085}$  & $\bm{0.082}$          \\ 

\bottomrule
\end{tabular}
}
\end{sc}
\end{small}
\end{center}
\end{table}

\section{Discussion}
\subsection{Computation Overhead}
We conduct the computation overhead experiments on NVIDIA RTX 4090 24GB GPUs. We use a single GPU with a batch size of 1 to optimize the camouflage (the one used in our paper was optimized for 4 epochs), with a GPU memory usage of 7.8 GB and an average epoch time of 221 minutes, compared to 203 minutes per epoch for the baseline method FCA. Breaking this down, environment feature extraction averages 3 minutes, ROA component takes 10 minutes, object detection requires 7 minutes, computing the loss and backpropagating gradients take 16 minutes, E2E-NR combines for 173 minutes, and the other remaining steps accounting for 11 minutes.

Our general end-to-end weight training is conducted on 4 GPUs, with a batch size of 4 per GPU. Each GPU consumes 18 GB of memory, and the total training time is 22 hours. Critically, finetuning these pre-trained E2E weights on specific models is far more efficient, typically requiring only an average of 10 minutes.

\subsection{Limitation}
The proposed camouflage generation framework is primarily designed to maximize attack effectiveness and robustness. However, it does not consider the naturalness of the appearance, leading to highly saturated colors and abrupt texture variations. Although this enhances adversarial strength, it also makes the camouflage more conspicuous to human observers. Moreover, the current approach targets object detectors using static images. In practice, vehicles operate in dynamic scenes and are often tracked by target tracking systems, where temporal factors such as motion blur and tracking dynamics may further reduce the effectiveness of the camouflage. These temporal dynamics are not accounted for in the present framework design.

\subsection{Future work}
We will focus on addressing the limitations identified in this study. To enhance the naturalness of the generated camouflage, integrating diffusion models presents a promising direction, as these models excel at producing high-fidelity and realistic images, which could reduce the visual conspicuousness while preserving attack effectiveness. Additionally, since real-world vehicles operate in dynamic scenes and are often tracked by target tracking systems, future work will focus on attacking these trackers by explicitly incorporating temporal information into the camouflage generation process, aiming to enhance the attack effectiveness over time and under motion-induced challenges such as motion blur.

\section{Conclusion}
We have proposed RAUCA, a robust and accurate UV-map-based physical adversarial camouflage attack framework with a realistic end-to-end neural renderer and a multi-weather dataset. In particular, we propose a novel neural render component, namely E2E-NRP, which offers the advantages of precise optimization of the UV map and the ability to render environmental characteristics. Additionally, we incorporate a multi-weather dataset during camouflage generation to further enhance its robustness. Moreover, we enhance the adaptability of the framework to different vehicles by enhancing the EFE training methodology with faster training speed and providing pre-trained EFE trained on multiple objects. Our experiments demonstrate that RAUCA outperforms the existing works under multi-weather situations, making it more robust in simulation and physical world settings.

\section*{Acknowledgements}

This work is supported by the National Natural Science Foundation of China (Grant No. 62306093), the Guangdong Provincial Key Laboratory of Novel Security Intelligence Technologies (Grant No. 2022B1212010005), and Shenzhen Science and Technology Program (Grant No. JSGGKQTD20221101115655027). The remaining support comes from the National Natural Science Foundation of China (Grant: 62376074), the National Key R\&D Program of China (Grant: 2021YFB2700900), and Shenzhen Science and Technology Program (Grants: KCXST20221021111404010, JSGG20220831103400002, SGDX20230116091244004, KJZD20230923114405011, KJZD20231023095959002, RKX20231110090859012), the Fundamental Research Funds for the Central Universities (Grants: HIT.DZJJ.2023118, HIT.OCEF.2024047), and the Fok Ying Tung Education Foundation of China (Grant: 171058).

\bibliographystyle{IEEEtran}
\bibliography{IEEEabrv,RAUCA_plus}

\newpage

\section{Biography Section}
 




\vskip 0pt plus -1fil
\begin{IEEEbiography}[{\includegraphics[width=1in,height=1.25in, clip,keepaspectratio]{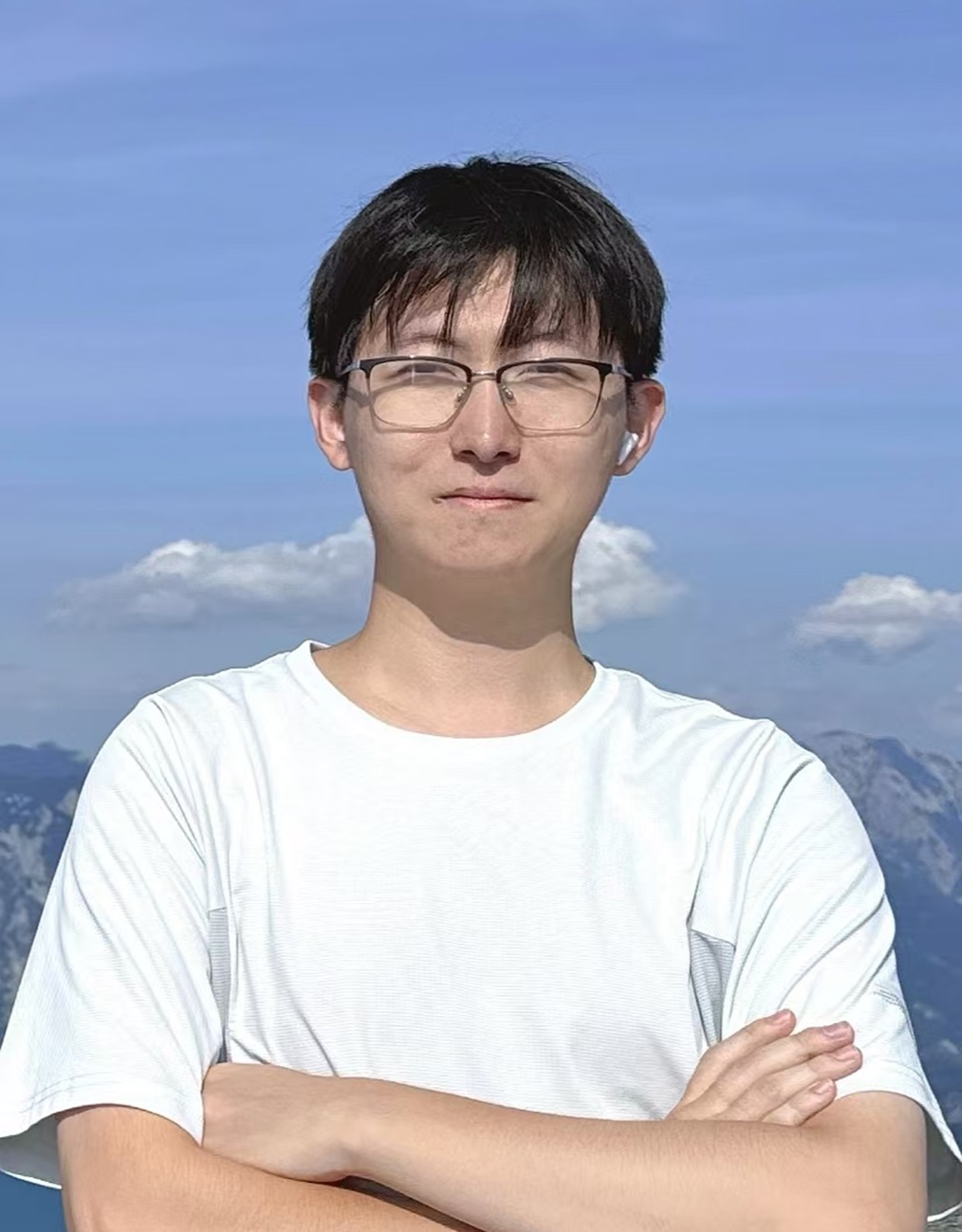}}]{Jiawei Zhou} is currently a postgraduate student major in Computer Science and Technology in Harbin Institute of Technology, Shenzhen. His research interest are the AI safety issues. He has published papers at leading conferences like ICML and NeurIPS. He obtained his bachelor’s degree from  Harbin Institute of Technology, Shenzhen in 2024.
\end{IEEEbiography}

\vskip 0pt plus -1fil
\begin{IEEEbiography}[{\includegraphics[width=1in,height=1.4in,clip,keepaspectratio]{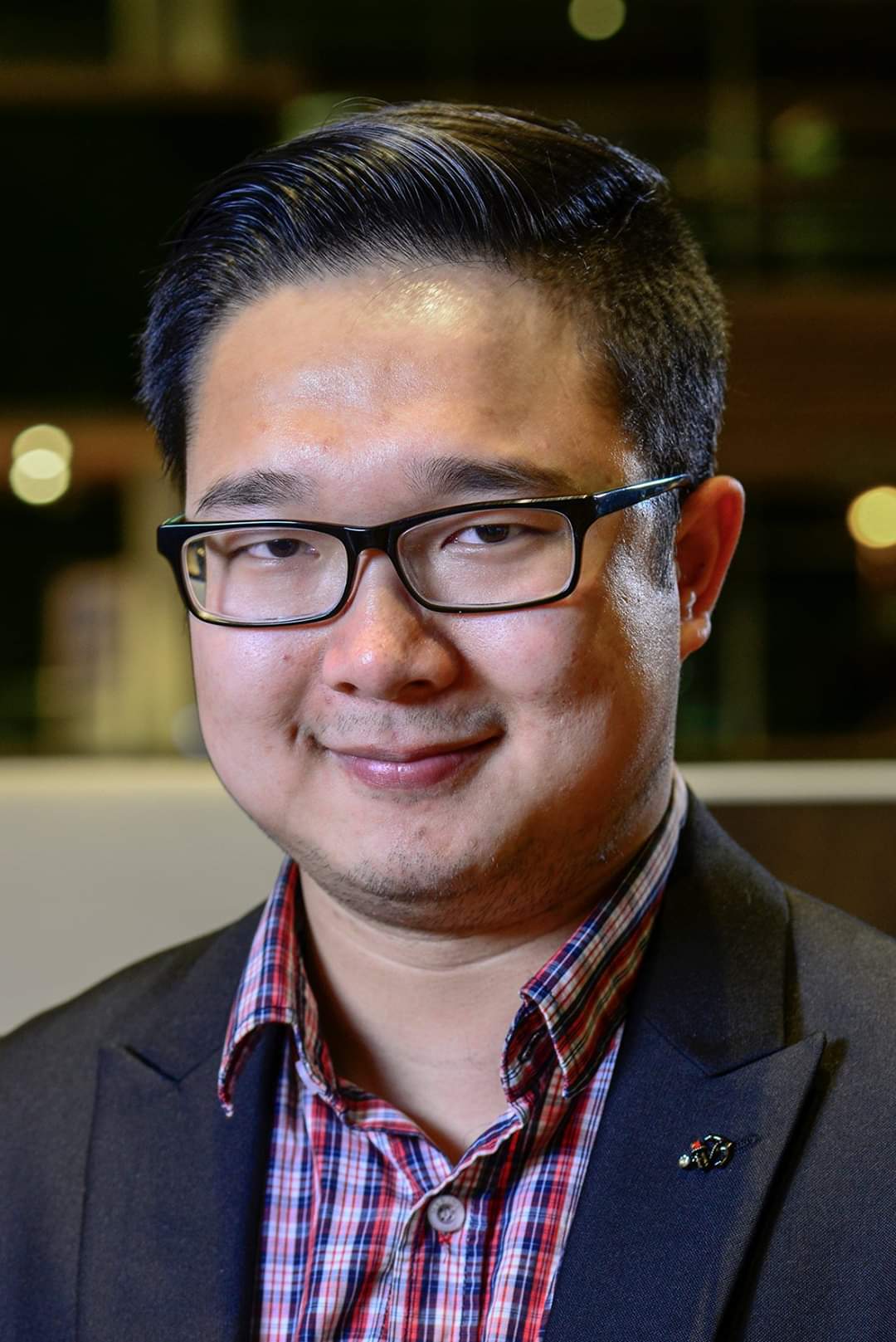}}]{Linye Lyu} is currently a Ph.D. student major in Computer Science and Technology in Harbin Institute of Technology, Shenzhen. His research interest are the AI safety issues in real-world scenarios such as autonomous driving and large languange models. He obtained two master degrees in Artificial Intelgience and Electronics and ICT Engineering from Katholieke Universiteit Leuven (KU Leuven) in 2021 and 2017, respectively. He obtained his bachelor’s degree from  KU Leuven and University
of Electronic Science and Technology of Chinea (UESTC) in 2016.
\end{IEEEbiography}

\vskip 0pt plus -1fil
\begin{IEEEbiography}[{\includegraphics[width=1in,height=1.25in, clip,keepaspectratio]{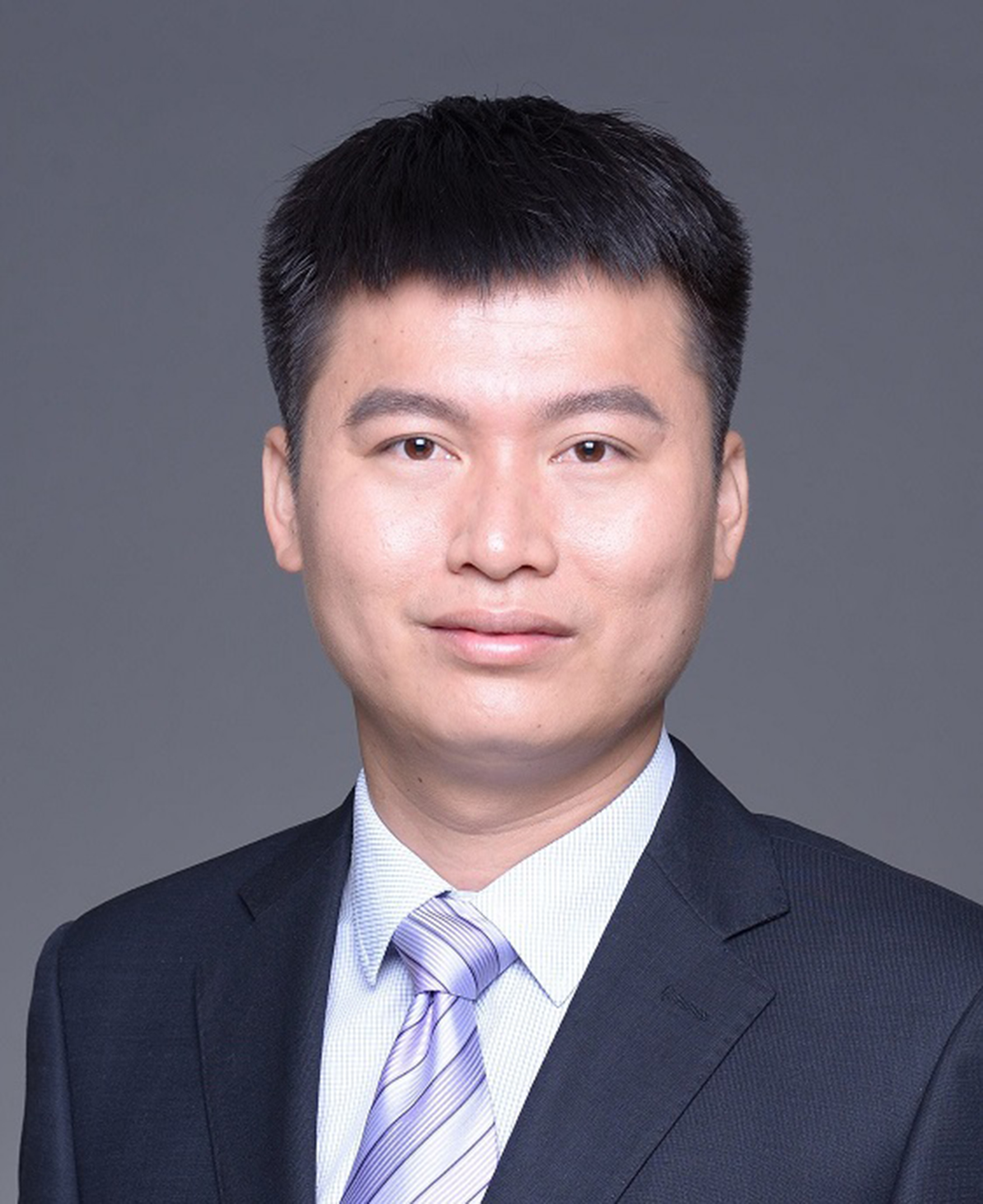}}]{Daojing He} received the B.Eng.(2007) and M. Eng. (2009) degrees from Harbin Institute of Technology (China) and the Ph.D. degree (2012) from Zhejiang University (China), all in Computer Science. He is currently a professor at the School of Computer Science and Technology, Harbin Institute of Technology, Shenzhen, China. His research interests include network and systems security. He is on the editorial board of some international journals such as IEEE Communications Magazine.
\end{IEEEbiography}

\vskip 0pt plus -1fil
\begin{IEEEbiography}[{\includegraphics[width=1in,height=1.5in,clip,keepaspectratio]{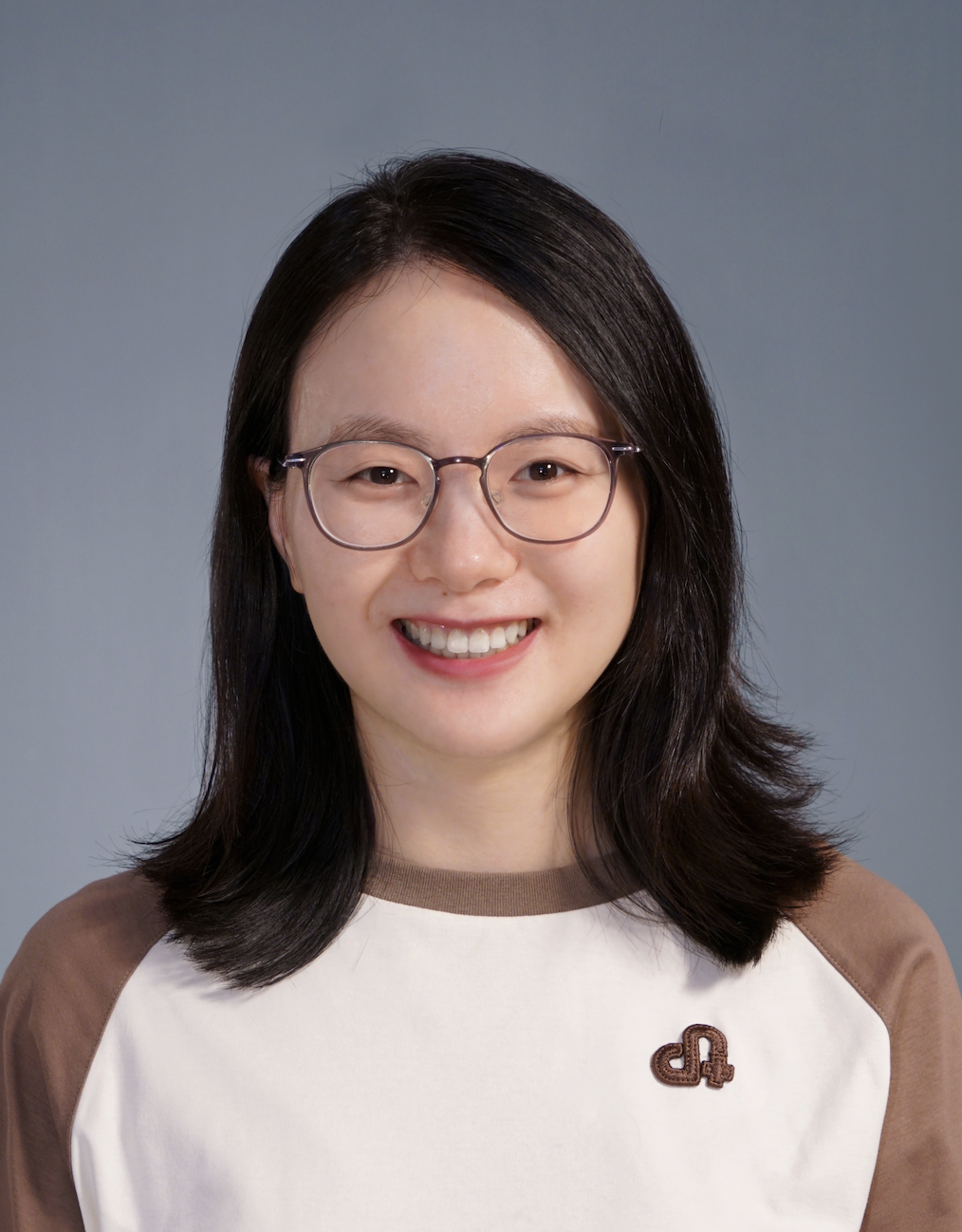}}]{Yu Li} is currently serving as a ZJU100 Professor at the School of Integrated Circuits, Zhejiang University. She received her Ph.D. degree from the Department of Computer Science and Engineering at the Chinese University of Hong Kong in 2022, obtained her master’s degree from the Katholieke Universiteit Leuven (KU Leuven) in 2017, and obtained her bachelor’s degree from the KU Leuven and the University
of Electronic Science and Technology (UESTC) in
2016. Li Yu’s main research direction is AI security
and testing. She has published many papers at leading conferences like CCS, NDSS, NeurIPS, ICML, and ISSTA.
She was nominated as a candidate for the 2022 Young Scholar Ph.D. Dissertation Award at The Chinese University of Hong Kong. In recognition of her outstanding work, she received the Best Ph.D. Dissertation Award at the 2022 Asian Test Symposium and the runner-up of the E. J. McCluskey Ph.D. Dissertation Award by the IEEE Test Technology Technical Council (TTTC).
\end{IEEEbiography}

\vfill

\end{document}